\documentclass{article} 
\usepackage{iclr2026_conference,times}

\usepackage{amsmath,amsfonts,bm}

\def\eqref#1{equation~\ref{#1}}

\def\1{\bm{1}}

\DeclareMathAlphabet{\mathsfit}{\encodingdefault}{\sfdefault}{m}{sl}
\SetMathAlphabet{\mathsfit}{bold}{\encodingdefault}{\sfdefault}{bx}{n}

\usepackage{hyperref}
\usepackage{url}

\usepackage[utf8]{inputenc} 
\usepackage[T1]{fontenc}    
\usepackage{booktabs}       
\usepackage{amsfonts}       
\usepackage{nicefrac}       
\usepackage{microtype}      
\usepackage{xcolor}         
\usepackage{amsmath}
\usepackage{graphicx}
\usepackage{subcaption}
\usepackage{float}         
\usepackage{caption}       
\usepackage{array}
\usepackage{wrapfig}
\usepackage{tabularx}
\usepackage{makecell}
\usepackage{marvosym}

\newtheorem{definition}{} 
\newtheorem{theorem}{}  
\newtheorem{proposition}{}  

\usepackage{booktabs}
\usepackage{multirow}
\usepackage{siunitx}
\title{Rate-Distortion Optimized Communication for Collaborative Perception}

\author{Genjia Liu\textsuperscript{1}, Anning Hu\textsuperscript{1}, Yue Hu\textsuperscript{2}, Wenjun Zhang\textsuperscript{1}, Siheng Chen\textsuperscript{1}\thanks{\Letter~Corresponding author} \\
\textsuperscript{1}Shanghai Jiao Tong University,
\textsuperscript{2} University of Michigan\\
\textsuperscript{1}\texttt{\{LGJ1zed,huanning,zhangwenjun,sihengc\}@sjtu.edu.cn} \\
\textsuperscript{2}\texttt{huyu@umich.edu}\\
}

\iclrfinalcopy 
\begin{document}

\maketitle

\begin{abstract}

Collaborative perception emphasizes enhancing environmental understanding by enabling multiple agents to share visual information with limited bandwidth resources. While prior work has explored the empirical trade-off between task performance and communication volume, a significant gap remains in the theoretical foundation. 
To fill this gap, we draw on information theory and introduce a pragmatic rate-distortion theory for multi-agent collaboration, specifically formulated to analyze performance-communication trade-off in goal-oriented multi-agent systems. This theory concretizes two key conditions for designing optimal communication strategies: supplying pragmatically relevant information and transmitting redundancy-less messages.
Guided by these two conditions, we propose RDcomm, a communication-efficient collaborative perception framework that introduces two key innovations: i) task entropy discrete coding, which assigns features with task-relevant codeword-lengths to maximize the efficiency in supplying pragmatic information;
ii) mutual-information-driven message selection, which utilizes mutual information neural estimation to approach the optimal redundancy-less condition.
Experiments on 3D object detection and BEV segmentation demonstrate that RDcomm achieves state-of-the-art accuracy on DAIR-V2X and OPV2V, while reducing communication volume by up to 108$\times$. The code will be released.

\end{abstract}


\section{Introduction}

Multi-agent collaborative perception enhances environmental understanding by enabling agents to jointly perceive and share information. This paradigm has shown clear advantages over single-agent sensing, particularly in overcoming occlusions and limited fields of view, and has been widely adopted in tasks such as 3D object detection~\cite{wang2020v2vnet} and BEV segmentation~\cite{XuCoBEVT:CoRL22}.

In this field emphasizing multi-agent collaboration, a fundamental challenge remains: the trade-off between task performance and communication volume~\cite{hu2022where2comm}. While sharing richer information tends to preserve collaboration quality, it introduces significant communication overhead; conversely, aggressively limiting communication may result in the loss of task-critical information, ultimately degrading overall performance.
Recent works tackle this trade-off by extracting informative and compact representations from visual observations to serve as collaborative messages. One representative line of work focuses on spatial selection, aiming to transmit only task-relevant regions, such as those with high detection confidence~\cite{hu2022where2comm} or sparse observation coverage~\cite{xu2025cosdh}. Another direction leverages neural compression techniques~\cite{balle2018variational,van2017vqvae}, reducing the size of transmitted features through autoencoders~\cite{shao2024task,hu2024codefilling} or channel reduction~\cite{li2021disconet, lu2024HEAL}.
Despite some empirical gains, previous approaches are heuristic in nature, relying on manually designed communication strategies or intuitive criteria. Such approaches lack theoretical grounding and provide no principled guidance on what to communicate or how to encode it under bandwidth constraints.

To fill this gap, we take an information-theoretic perspective and propose \textit{pragmatic rate-distortion theory for multi-agent collaboration}, which explicitly models the trade-off between communication bit-rate and task-specific pragmatic distortion.
Our theoretic analysis extends Shannon’s classical rate-distortion framework~\cite{shannon1959coding} in two key aspects. First, we introduces pragmatic distortion, a task-driven metric that reflects the impact of message degradation on downstream task performance, distinct from reconstruction-based distortions~\cite{blau2019rethinking_distortion,cover1999elements_infor_theory}. Second, our theory generalizes to distributed communication among multiple agents, where both message senders and receivers observe the environment. We thus account for inter-agent redundancy, a factor typically neglected in traditional rate–distortion analysis.
Building upon these extensions, our theory ultimately characterizes the minimal communication cost required to meet a specified distortion threshold, and derives two key conditions that an optimal communication strategy should satisfy: \textit{pragmatic-relevant} and \textit{redundancy-less}. We envision this theoretical framework as a foundation for analyzing communication efficiency in broader multi-agent tasks.

Inspired by the two theoretical conditions, we propose \textbf{RDcomm} (\textbf{R}ate-\textbf{D}istortion guided pragmatic \textbf{comm}unication), a novel communication-efficient collaborative perception system, which 
optimizes both message selection and coding to reduce communication overhead while preserving collaborative complementarity and task effectiveness. Specifically, we design the two core components of RDcomm based on the two derived conditions: i) Based on the pragmatic-relevant condition, we propose a novel task entropy discrete coding module. It first utilizes learned codebooks to quantize feature vectors, and then applies variable-length coding guided by task relevance, assigning shorter codewords to more informative features. 
ii) Based on the redundancy-less condition, we propose a novel feature selection module leveraging mutual information neural estimation~\cite{belghazi2018MINE}. This module enables agents to perform an inter-agent handshake process to assess message redundancy by quantifying the mutual information between shared and locally observed features.
We validate RDcomm on two representative perception tasks: 3D object detection and BEV semantic segmentation, using both real-world dataset DAIR-V2X~\cite{YuDAIRV2X:CVPR22} and simulation dataset OPV2V~\cite{XuOPV2V:ICRA22}. Experimental results show that RDcomm reduces communication volume by up to 108 times compared against existing methods.

Our main contributions are summarized as follows:
\begin{itemize}
\vspace{-2mm}
    \item We introduce a pragmatic rate–distortion theory for multi-agent collaboration, which characterizes the performance–communication trade-off, and concretize two optimal conditions: i) supply pragmatic-relevant information; ii) avoid inter-agent redundancy.
    \vspace{-1mm}
    \item We propose RDcomm, a communication-efficient collaborative perception framework that is designed to approach the two optimal conditions with two innovations: i) task entropy discrete coding; ii) mutual-information-driven message selection. Experiments on detection and segmentation tasks demonstrate that RDcomm achieves dual superiority in both performance and communication efficiency.
\end{itemize}

\vspace{-2mm}
\section{Related works}
\vspace{-2mm}
\subsection{Communication-efficient collaborative perception}
\vspace{-2mm}

In multi-agent collaboration, a key challenge is to balance task performance and communication cost~\cite{hu2022where2comm}. 
Early collaboration transmits raw sensor data~\cite{han2023survey} and achieves high accuracy but suffers from heavy bandwidth usage. 
Late collaboration reduces bandwidth by sending final predictions, but degrades performance under noise~\cite{lu2023robust,hu2022where2comm}.
To address this, intermediate collaboration transmits feature maps to strike a balance between performance and efficiency. Prior works mainly improve efficiency via: i) spatial selection, which transmits features at critical regions; and ii) feature compression.
For spatial selection, Where2comm~\cite{hu2022where2comm} selects high-confidence regions, CodeFilling~\cite{hu2024codefilling} removes redundant collaborators, and CoSDH~\cite{xu2025cosdh} targets unobserved areas. For compression, techniques include value quantization~\cite{wang2020v2vnet, shao2024task}, vector quantization~\cite{hu2024codefilling}, and channel reduction~\cite{li2021disconet, lu2024HEAL}.
However, these methods are primarily heuristic and lack theoretical guarantees for communication efficiency. In our work, we provide a theoretical framework grounded in rate-distortion analysis, offering explicit conditions for optimal communication.
\vspace{-2mm}
\subsection{Rate-distortion theory background}
\vspace{-2mm}
Rate-distortion theory~\cite{shannon1959coding} provides a fundamental framework for lossy compression by characterizing the minimum bits required to represent a signal $X$ as a compressed representation $Z$ under a specified distortion constraint $\mathrm{D}[X, Z] \leq \delta$. The goal is to find a probabilistic encoding map $p(Z|X)$ that minimizes the mutual information $\mathrm{I}(X; Z)$ , as formulated in~(\ref{eq:RD_shannon}):
\begin{align}
    \operatorname{Rate}(\delta) = \min_{p(Z|X)} \mathrm{I}(X; Z) \quad \mathrm{s.t.} ~ \mathrm{D}[X, Z] \leq \delta
    \label{eq:RD_shannon}
\end{align}
In lossy compression, $Z$ is an approximate reconstruction of $X$. A typical example is a Gaussian source $X \sim \mathcal{N}(\mu, \sigma^2)$ with mean squared error (MSE) distortion, where the optimal rate under distortion level $\delta$ has a closed-form solution $R(\delta) = h(X) - \frac{1}{2} \log (2\pi e \delta)$, which reflects the total information in $X$ minus the portion tolerable under the distortion budget. 
Despite its general application in visual compression~\cite{balle2018variational}, classical rate-distortion analysis mostly caters to fidelity-based distortion metrics~\cite{blau2019rethinking_distortion} and single-source settings. In general, any distortion measure $d: \mathcal{X} \times \mathcal{Z} \to \mathbb{R}_{\geq 0}$ of the form $D[X,Z] = \mathbb{E}_{p(x,z)}[d(x,z)]$ is valid, as long as there exists a $z \in \mathcal{Z}$ such that $D[X,Z]$ is finite~\cite{dubois2021lossy}. Our work extends this framework to multi-agent collaborative perception by incorporating task-specific distortion and inter-agent redundancy. 

\section{Pragmatic rate-distortion
theory for collaboration}
We introduce the pragmatic rate-distortion theory for multi-agent collaboration. Our theoretical analysis follows three high-level steps:
i) We reformulate the collaboration task objective in a rate-distortion formulation (Sec.~\ref{sec:problem_formulation});
ii) Defining the pragmatic distortion for collaboration (Sec.~\ref{sec:prag_distortion});
iii) We derive the minimal transmission rate under constrained distortion, and present the conditions for an optimal communication strategy (Sec.~\ref{sec:optimal_rate}). See detailed proofs in the Appendix~\ref{sec:proofs}.

\vspace{-2mm}
\subsection{Problem formulation}
\vspace{-2mm}
\label{sec:problem_formulation}
We first introduce the problem we target to solve in rate-distortion formulation. 
In collaborative tasks, our goal is to optimize model parameters and the message generation strategy, in order to achieve the minimal transmission bits under constrained task loss, that is~(\ref{eq:collaboration_define}): 
\vspace{-1mm}
\begin{align}
\min_{Z_s,\theta}\sum_{r=1}^N  |\{Z_{s\to r}\}_{s\neq r}| \quad \text{s.t.} ~  \sum_{r=1}^N L_Y\left(Y_r,\Phi_\theta\left(X_r, \{Z_{s\to r}\}_{s\neq r}\right) \right) \leq L_{max}
\label{eq:collaboration_define}
\vspace{-2mm}
\end{align}
where $N$ denotes the number of agents; $\{Z_{s \to r}\}_{s \neq r}$ represents the messages sent from agent $s$ to other agents; $|\cdot|$ measures the information volume; $\Phi_{\theta}(\cdot)$ is the task model parameterized by $\theta$; $X_r$ and $Y_r$ denote the local observation and ground-truth label of agent $r$, respectively; $L_Y(\cdot)$ is the loss function associated with task $Y$, and $L_{\max}$ specifies the maximum tolerable task loss. We then present~(\ref{eq:collaboration_define_RD})  as a formal rate-distortion optimization version of~(\ref{eq:collaboration_define}), which defines the minimal communication bits $\operatorname{Rate}(\delta)$ and serves as the foundational objective.
\begin{align}
\operatorname{Rate}(\delta)=\min _{p(Z_{s\to r} \mid X_s)} \mathrm{I}(X_s ; Z_{s\to r}) \quad \text { s.t. } \mathrm{D}_Y[X_s, Z_{s\to r}|X_r] \leq \delta .
\label{eq:collaboration_define_RD}
\end{align}
As shown in~(\ref{eq:collaboration_define_RD}), the communication volume is captured by $\mathrm{I}(X_s ; Z_{s\to r})$, which quantifies the amount of information from the original observation $X_s$ that is preserved in the transmitted message $Z_{s\to r}$.
We denote $\mathrm{D}_Y[X_s, Z_{s\to r} \mid X_r]$ as the pragmatic distortion for the collaborative task $Y$, which measures the degradation in task performance when transmitting $Z_{s\to r}$ instead of $X_s$, given the local observation $X_r$. See detailed discussions on our problem formulation in Appendix~\ref{sec:discuss_problem_formulation}.



\vspace{-2mm}
\subsection{Pragmatic distortion for collaborative perception}
\vspace{-2mm}
\label{sec:prag_distortion}
In this section we make the pragmatic distortion $\mathrm{D}_Y[X_s, Z_{s\to r} \mid X_r]$ in~(\ref{eq:collaboration_define_RD}) explicit for collaborative perception.
Let $X_s$, $X_r$ denote the sender’s and receiver’s local observations, respectively, $Y$ denotes the perception task target. $Z_{s \rightarrow r}$ is compressed from $X_s$.
We define 
$\mathrm{D}_Y[X_s, Z_{s \rightarrow r} | X_r]$ in~(\ref{eq:pragmatic_distortion}):
\begin{align}
\mathrm{D}_Y\left[X_s, Z_{s\rightarrow r}|X_r\right]=\mathrm{B}_{risk}\left[Y | Z_{s\rightarrow r},X_r\right]-\mathrm{B}_{risk}\left[Y | X_s,X_r\right]
\label{eq:pragmatic_distortion}
\end{align}
where $\mathrm{B}_{\mathrm{risk}}[Y|X_r, X_s]$ denotes the Bayes Risk~\cite{dubois2021lossy}, which measures the minimum achievable prediction error for the target $Y$ given the joint inputs $X_r$ and $X_s$. 
We define the distortion $\mathrm{D}_Y[X_s, Z_{s\rightarrow r} | X_r]$ as the increase in Bayes Risk when predicting $Y$ with the compressed representation $Z_{s\rightarrow r}$ instead of the original signal $X_s$, while conditioning on the existing local information $X_r$. 
Formally, the Bayes Risk is
$\mathrm{B}_{\mathrm{risk}}[Y | X] 
= \inf_{f} \; \mathbb{E}_{p(X,Y)} \left[ L\big(Y, f(X)\big) \right]$,
where $f$ is any predictor.  
For perception tasks, the loss is typically computed independently at each BEV location.
Accordingly, we define the overall Bayes Risk as the average of pixel-wise Bayes Risks over all locations, as in~(\ref{eq:bayes_risk_perception}),
where $i \in \mathcal{S}$ denotes a BEV coordinate within the perception range $\mathcal{S}$, 
and $Y_{(i)}$ and $f(X)_{(i)}$ denote the corresponding ground-truth label and model prediction.
\begin{align}
    \mathrm{B}_{risk}[Y|X] = \inf_{f} \mathbb{E}_{p(X, Y)} \left[ L(Y, f(X)) \right]=  \inf_{f} \frac{1}{|Y|}\sum_{i\in \mathcal{S}} ~ \mathbb{E}_{p(X, Y_{(i)})} \left[ L(Y_{(i)}, f(X)_{(i)}) \right]
    \label{eq:bayes_risk_perception}
\end{align}
We instantiate the pragmatic distortion for two representative perception tasks: BEV segmentation and 3D object detection. Specifically, for BEV segmentation, we adopt the per-pixel cross-entropy (CE) loss; for 3D object detection, we adopt the widely used CenterPoint loss~\cite{yin2021center}.
We directly present the final derivations of the pragmatic distortion \(\mathrm{D}_Y\left[X_s, Z_{s\rightarrow r} | X_r\right]\) in Tab.~\ref{tab:distortions}. 


\vspace{-2mm}
\begin{table}[h]
\caption{\small Pragmatic distortions differ from classical reconstruction distortion by considering task entropy $\mathrm{H}(Y|\cdot)$ and local redundancy $X_r$. Proofs are provided in Appendix~\ref{proof:distor_coperception}.}
\vspace{-2mm}
\centering
\label{tab:distortions}
\resizebox{\linewidth}{!}{
\begin{tabular}{lc}
\toprule
\textbf{Task (loss function)} & \textbf{Distortion $\mathrm{D}_Y\left[X_s, Z_{s\rightarrow r}|X_r\right]$} \\ \midrule
Lossy reconstruction~\cite{balle2018variational} (MSE) & 
\begin{minipage}{\linewidth}
\begin{equation}
\small
\frac{1}{|X_s|} \left\| X_s - Z_{s\rightarrow r} \right\|_2^2, \quad \text{no}~X_r
\label{eq:distortion_recons}
\end{equation}
\end{minipage}
\\
\textbf{BEV segmentation (CE)}  & 
\begin{minipage}{\linewidth}
\begin{equation}
\small
\frac{1}{|Y|} 
\sum_{i \in \mathcal{S}} [ 
    \mathrm{H}\!\left( Y_{(i)} \mid Z_{s\to r}, X_r \right) 
    - \mathrm{H}\!\left( Y_{(i)} \mid X_s, X_r \right)
]
\label{eq:distortion_seg}
\end{equation}
\end{minipage}
\vspace{-3mm}
\\
\textbf{3D detection. (CenterPoint)}  & 
\begin{minipage}{\linewidth}
\begin{equation}
\small
\frac{1}{|Y|} 
\sum_{i \in \mathcal{S}} [ 
    \mathrm{H}\!\left( Y_{(i,c)} \mid Z_{s\to r}, X_r \right) 
    - \mathrm{H}\!\left( Y_{(i,c)} \mid X_s, X_r \right) 
    + \frac{1}{2} 
    \sum_{k \in \mathcal{K}} \left( 
        e^{ \mathrm{H}( Y_{(i,k)} \mid Z_{s\to r}, X_r ) - 1 }
        - e^{ \mathrm{H}( Y_{(i,k)} \mid X_s, X_r ) - 1 }
    \right) 
]
\label{eq:distortion_det}
\end{equation}
\end{minipage}
\\ \bottomrule
\end{tabular}
}
\end{table}
\vspace{-2mm}

\subsection{Minimal bit-rate of collaborative message and optimal conditions}
\vspace{-2mm}

In this section, we present the trade-off between communication bit-rate and distortion by incorporating the pragmatic distortions~(\ref{eq:distortion_seg}) and~(\ref{eq:distortion_det}) into objective~(\ref{eq:collaboration_define_RD}). 
\begin{theorem}
    \label{theor:optimal_rate_collab}
(Minimal bit-rate $\operatorname{Rate}(\delta)$ of collaborative message under distortion constraint, see proof in~\ref{proof:collab_optimal_rate}).
Consider a message sender agent $a_s$ and a message receiver agent $a_r$ and their observation denoted as $X_s,X_r$, where the sender compresses $X_s$ as $Z_{s\rightarrow r}$ and transmits it to the receiver to collaborates in achieving task target $Y$. Then, the minimal transmission bit-rate $\operatorname{Rate}(\delta)=\min _{p(Z_{s\rightarrow r} \mid X_s)} \mathrm{I}(X_s ; Z_{s\rightarrow r})$ s.t. $\mathrm{D}_Y[X_s, Z_{s\rightarrow r}] \leq \delta$ is given in~(\ref{eq:rate_distortion_collab_final_0}). 
\begin{align}
\operatorname{Rate}(\delta)
=&\min _{p(Z_{s\rightarrow r} \mid X_s)~\text { s.t. } \mathrm{D}_Y[X_s, Z_{s\rightarrow r}] \leq \delta } \mathrm{I}(X_s ; Z_{s\rightarrow r})=\mathrm{I}(Y;X_s \mid X_r)-\delta
\\
=& \mathrm{H}(X_s) - \underbrace{ \mathrm{H}(X_s|Y) }_{\text{information in }X_s\text{ irrelevant to }Y} - \underbrace{ \mathrm{I}(Y;X_s;X_r)}_{\text{information in }X_s\text{ redundant with }X_r\text{ about }Y} -\delta \hfill
\label{eq:rate_distortion_collab_final_0}
\end{align}
\end{theorem}

The minimal bit-rate $\operatorname{Rate}(\delta)$ can be achieved only if the following two conditions are satisfied.
\vspace{-0.5mm}
\textbf{Pragmatic-relevant.} The transmitted message $Z_{s\rightarrow r}$ should contain only information relevant to the receiver’s task $Y$, as formalized in~(\ref{eq:task-efficient}): 
\vspace{-2mm}
\begin{align}
    \mathrm{H}(Z_{s\rightarrow r}|Y)=0
    \label{eq:task-efficient}
\end{align}
Equation~(\ref{eq:task-efficient}) implies that the uncertainty of $Z_{s\rightarrow r}$ is eliminated given a specific task target $Y$, indicating a unique mapping from $Y$ to $Z_{s\rightarrow r}$. Consequently, $Z_{s\rightarrow r}$ should exclude any information unrelated to the task $Y$, and the task-relevant messages should be prioritized during coding.

\textbf{Redundancy-less.} The transmitted message $Z_{s\rightarrow r}$ should avoid maintaining information that is already contained in the receiver's observation $X_r$,  which is expressed as~(\ref{eq:Supply-exact}):
\begin{align}
    \mathrm{I}(Z_{s\rightarrow r};X_r)=0
    \label{eq:Supply-exact}
\end{align}
Equation~(\ref{eq:Supply-exact}) shows that the mutual information between the transmitted message $Z_{s\rightarrow r}$ and the observation of receiver $X_r$ should be eliminated. In other words, $Z_{s\rightarrow r}$ should avoid containing the redundant information, but exactly supply the information missed in $X_r$. 



\label{sec:optimal_rate}
\section{RDcomm: efficient communication for co-perception}
\label{sec:rdcomm}

\begin{figure}[!t]
\centering
\includegraphics[width=0.97\linewidth]{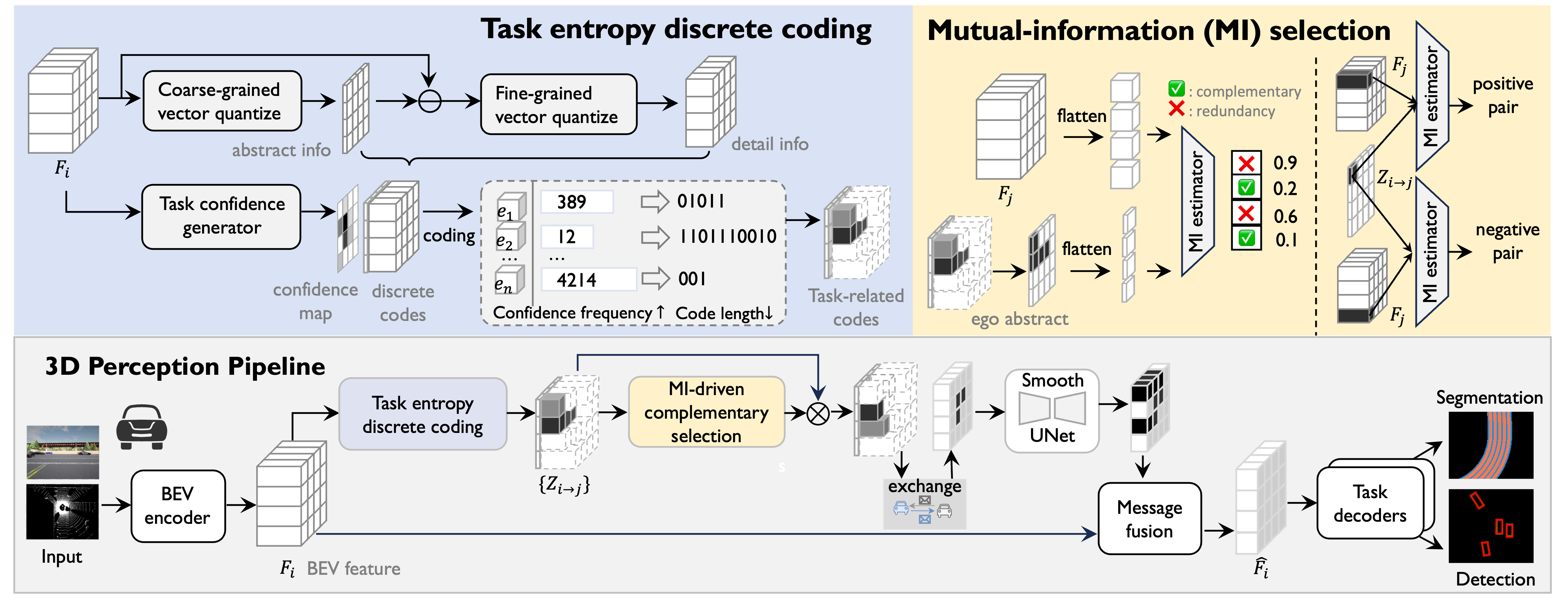}
\vspace{-1.5mm}
\caption{\small RDcomm features two key components: i) task entropy
discrete coding for improving the pragmatic relevance of message, which assigns short codewords to the codes with high confidence frequency; ii) mutual-information-driven message selection, which measures message redundancy by mutual information estimation.}
\label{fig:rdcomm_overview}
\vspace{-5mm}
\end{figure}

Inspired by the theoretical analysis, we introduce \textbf{RDcomm}, a novel communication-efficient collaborative perception framework with three main components (Fig.~\ref{fig:rdcomm_overview}):
i) a perception pipeline that provides the basic functionalities for perception tasks;  
ii) a task entropy discrete coding module (Sec.~\ref{sec:task_encoding}), following the pragmatic-relevant condition (\ref{eq:task-efficient}), which adopts a novel variable-length coding guided by task relevance;  
iii) a mutual-information–driven selection module (Sec.~\ref{sec:MI_selection}), following the redundancy-less condition (\ref{eq:Supply-exact}), which selects complementary messages for transmission.

\vspace{-1mm}
\textbf{Perception pipeline.} The perception pipeline couples a BEV encoder with task-specific decoders. The BEV encoder accepts either LiDAR or camera inputs and maps sensor data into a unified bird’s-eye-view (BEV) representation, enabling consistent spatial alignment across agents. Denoting the observation of the sender agent by $X_s$, the BEV encoder $\Phi_{\mathrm{bev}}(\cdot)$ produces a BEV feature map by $F_s=\Phi_{\mathrm{bev}}(X_s)\in\mathbb{R}^{h\times w\times c}$, followed by task-specific decoders $\Phi_{\mathrm{task}}$ for downstream tasks such as 3D object detection and BEV segmentation. Backbone details are provided in Appendix~\ref{sec:single-agent-perception}. We then focus on compressing the feature $F_s$ into a collaboration message $Z_{s\rightarrow r}$ for any receiver agent.


\vspace{-1mm}
\subsection{Task entropy discrete coding}
\label{sec:task_encoding}
\vspace{-1mm}

Our first objective is to approach the pragmatic-relevant condition $\mathrm{H}(Z_{s\rightarrow r}|Y)=0$ in (\ref{eq:task-efficient}). We approximate it by minimizing the task-conditioned entropy $\mathrm{H}(Z_{s\rightarrow r}|Y)$.


\vspace{-1mm}
\textbf{Layered vector quantization.} We begin by constraining $\mathrm{H}(Z_{s\rightarrow r})$ via vector quantization inspired by~\cite{zhu2022unified}, where the core idea is to represent each vector in $F_s$ with the nearest embedding $\mathbf{e}_{i}$ in a codebook $\mathbf{B}=[\mathbf{e}_{1},\mathbf{e}_{2},\dots,\mathbf{e}_{n}] \in \mathbb{R}^{n\times d}$ with $n$ learnable embeddings. We further implement a layered discrete auto-encoder $\Phi_{\mathrm{vq}}(\cdot)$ to quantize the BEV feature $F_s$ by $F_s^q=\Phi_{\mathrm{vq}}(F_s,\mathbf{B}_{base},\mathbf{B}_{res})$. 
$\mathbf{B}_{base}$ is used to approximate the basic coarse-grained information of $F_s$ with small codebook volume $n$, and the residual error $F_s-\Phi_{\mathrm{vq}}(F_s,\mathbf{B}_{base})$ is further approximated by a fine-grained codebook $\mathbf{B}_{res}$ with larger volume. This layered quantization is described in (\ref{eq:mult-layer-vq0})(\ref{eq:mult-layer-vq1}):
\begin{align}
\label{eq:mult-layer-vq0} 
    Z_{base}^q &= \arg\min_{i} \|\mathbf{B}_{base}[i] - f_{in}(F_s)\|_2, \quad
    Z_{res} = F_s - Z_{base}^q
    \\ Z_{res}^q &= \arg\min_{i} \|\mathbf{B}_{res}[i] - f_{in}(Z_{res})\|_2 ,\quad
    F_s^q =  f_{out}(Z_{res}^q + Z_{base}^q)
    \label{eq:mult-layer-vq1}
\end{align}

\vspace{-2mm}
where $f_{in}(\cdot),f_{out}(\cdot)$ are MLP projectors to bridge the distribution gap between continuous vectors and codebook embeddings, and the input feature map $F_s$ is flattened before quantization.



\textbf{Task-aware priority and encoding.}
While quantization reduces representation length by restricting the vector-space volume, we further improve coding efficiency by introducing task bias. Specifically, we prioritize task-relevant messages for selection and encode them with shorter code lengths. Recall that our objective is to minimize the task-conditioned entropy $\mathrm{H}(Z_{s\rightarrow r} | Y)\to 0$, which can be expanded as (\ref{eq:conditional-entropy-expand}).
\vspace{-2mm}
\begin{align}
    \min_{Z_{s\rightarrow r}} \mathbb{E}_Y\sum_{Z_{s\rightarrow r}} [- p(Z_{s\rightarrow r}|Y)\log p(Z_{s\rightarrow r}|Y)]
    \label{eq:conditional-entropy-expand}
\end{align}
Note that maximizing $p(Z_{s\rightarrow r}|Y)$ to 1 provides a sufficient solution for minimizing $\mathrm{H}(Z_{s\rightarrow r}|Y)$, and $p(Z_{s\rightarrow r}|Y)\propto p(Y|Z_{s\rightarrow r})p(Z_{s\rightarrow r})$ for a given target distribution $p(Y)$. Therefore we priority the messages with high task confidence $p(Y|Z_{s\rightarrow r})$ for transmission. 
We implement this with a confidence generator $\Phi_{\mathrm{conf}}(\cdot)$ producing scores $C_s=\Phi_{\mathrm{conf}}(F_s)\in\mathbb{R}^{h\times w}$. 
The confidence mask is $M_c=\mathbf{1}[C_s>\tau_c]$, where $\tau_c$ is a confidence threshold, and quantized features are selected as $F^{q}_{sc}=M_c\odot F_s^q$. 
In practice, we instantiate $\Phi_{\mathrm{conf}}(\cdot)$ by reusing the task decoder $\Phi_{\mathrm{task}}(\cdot)$.

\vspace{-1mm}
We further reduce the average coding length of the quantized $F^{q}_{sc}$, where we encode each embedding $\mathbf{e}_{i}$ in $\{\mathbf{B}_{base},\mathbf{B}_{res}\}$ considering the joint effect of task confidence $p(Y|\mathbf{e}_{i})$ and occurrence frequency $p(\mathbf{e}_{i})$. Specifically, we define the confidence frequency $p_c(\cdot)$ for each embedding $\mathbf{e}_{i}$ as (\ref{eq:confidence_frequency}):
\begin{align}
    p_c(\mathbf{e}_{i}) = \sum_{F_s} \sum_{\{(u,v):~\mathbf{e}_{i}\in \Phi_{\mathrm{vq}}(F_s)[u,v]\}} \Phi_{\mathrm{conf}}(F_s)[u,v]
    \label{eq:confidence_frequency}
\end{align}

where $p_c(\mathbf{e}_{i})$ represents the total task confidence predicted from $\mathbf{e}_{i}$ across the entire dataset. We compute it by accumulating the confidence scores $\Phi_{\mathrm{conf}}(F_s)[u,v]$ at spatial locations $(u,v)$ where $F_s[u,v]$ are quantized with embedding $\mathbf{e}_{i}$. 
To improve coding efficiency for the task-relevant embeddings, we propose to assign shorter code lengths for the embeddings with higher confidence frequency. For implementation, this work provides a straightforward yet effective solution by applying Huffman coding~\cite{huffman2007method}, where we set the Huffman weight of $\mathbf{e}_{i}$ to be its confidence frequency $p_c(\mathbf{e}_{i})$. Our task-entropy coding ultimately produces an index map $D_{s}\in (\{0,1\}^{l_{u,v}})^{h\times w}$ to represent the input feature $F_s\in \mathbb{R}^{h\times w\times c}$, where $\{0,1\}^{l_{u,v}}$ denotes binary strings of variable length ${l_{u,v}}$. $D_{s}[u,v]=[D^{base}_{s}[u,v] \,\|\, D^{res}_{s}[u,v]]\in\{0,1\}^{l_{u,v}}$ denotes the code at location $(u,v)$, which consists of coarse-grained semantic information $D_s^{base}[u,v]$ and fine-grained $D_s^{res}[u,v]$.





\textbf{Discussion.}
i) The relation between coding method and the information optimization target (\ref{eq:conditional-entropy-expand}): the high-confidence selection module $\Phi_{\mathrm{conf}(\cdot)}$ reduces the task-conditioned entropy $\mathrm{H}(Z_{s\to r}|Y)$ governed by $p(Z_{s\rightarrow r}|Y)\propto p(Y|Z_{s\rightarrow r})p(Z_{s\rightarrow r})$. We instantiate this by defining the confidence frequency $p_c(\cdot)$, and use it as the weight for entropy coding.
ii) The intuition behind is simple: with limited bandwidth, high-confidence messages are selected more often, making their embeddings frequent. Accordingly, we weight entropy coding using the confidence frequency $p_c(\mathbf{e}_{i})$. 
iii) We use the accumulation of confidence rather than the average to distinguish embeddings with similar task relevance but different occurrence frequencies.
iv) The confidence-frequency coding is applied post-training and remains lossless for task performance, as it only reassigns embedding indices.


\subsection{Complementary Selection with Mutual Information Estimation}
\label{sec:MI_selection}

In this section we focus on reducing inter-agent message redundancy according to the redundancy-less condition $\mathrm{I}(Z_{s\to r};X_r)=0$ (\ref{eq:Supply-exact}). We approximate this target by~(\ref{eq:minimize_MI}). $\Omega$ denotes the selection region.
\begin{align}
    \min_{\Omega} \mathrm{I}(\hat{F}^q_{sc}[\Omega];F_r)
    \label{eq:minimize_MI}
\end{align}
To obtain the redundancy-less region $\Omega$, we perform feature selection via mutual information neural estimation: we approximate $\mathrm{I}(\hat{F}^q_{sc};F_r)$ by with a learnable estimator $\Phi_{\mathrm{MI}}(\hat{F}^q_{sc},F_r)$ and select features accordingly.
Note that we use the coarse-grained compression $\hat{F}^q_{sc}=\mathbf{B}_{base}[D_s^{base}]$ as an abstract of $F_s$, which is pre-handed to the receiver to estimate its redundancy with the receiver’s information $F_r$. This abstraction $\hat{F}^q_{sc}$ helps estimate semantic redundancy by transmitting $D_s^{base}$ with much smaller communication volume, which is 10 times smaller than the lossless message $D_s$.


\vspace{-1mm}
\textbf{Mutual information neural estimation.}
Consider the variable pair $\mathbf{s},\mathbf{r}\in \mathbb{R}^c$ that are included in $\hat{F}^q_{sc},F_r\in\mathbb{R}^{h\times w\times c}$ respectively, the mutual information $\mathrm{I}(\mathbf{s},\mathbf{r})$ can be estimated using~(\ref{eq:MINE_origin}):
\begin{align}
    \mathrm{I}(\mathbf{s},\mathbf{r}):=D_{KL}(\mathbb{P}_{\mathbf{s},\mathbf{r}}\parallel\mathbb{P}_\mathbf{s}\otimes\mathbb{P}_\mathbf{r}) 
    \geq \sup_{T:\mathbb{R}^c\times\mathbb{R}^c\to\mathbb{R}}\left\{\mathbb{E}_{\mathbb{P}_{\mathbf{s},\mathbf{r}}}[T(\mathbf{s},\mathbf{r})]-\log(\mathbb{E}_{\mathbb{P}_{\mathbf{s}}\otimes\mathbb{P}_{\mathbf{r}}}[e^{T(\mathbf{s},\mathbf{r})}])\right\}
    \label{eq:MINE_origin}
\end{align}
where $\mathbb{P}_{\mathbf{s},\mathbf{r}}$ denotes the joint distribution of $\mathbf{s}$ and $\mathbf{r}$, and $\mathbb{P}_\mathbf{s},\mathbb{P}_\mathbf{r}$ denote marginals; $D_{KL}$ is KL divergence; $T:\mathbb{R}^c\times\mathbb{R}^c \rightarrow \mathbb{R}$ is any function that projects the $(\mathbf{s},\mathbf{r})$ pair into a real number. Formulation~(\ref{eq:MINE_origin}) shows that mutual information $\mathrm{I}(\mathbf{s},\mathbf{r})$ actually measures the divergence between the joint distribution $\mathbb{P}_{\mathbf{s},\mathbf{r}}$ and marginal distribution $\mathbb{P}_\mathbf{s}\otimes\mathbb{P}_\mathbf{r}$. While (\ref{eq:MINE_origin}) employs the KL divergence as a standard estimation, we instead adopt the GAN-style divergence~\cite{nowozin2016fgan} following~\cite{li2020gxn} to facilitate optimization, and use the estimation target in (\ref{eq:MINE_gan}). $\sigma(\cdot)$ denotes sigmoid.
\begin{align}
    \hat{\mathrm{I}}(\mathbf{s},\mathbf{r}) \geq \sup_{T:\mathbb{R}^c\times\mathbb{R}^c\to\mathbb{R}}\left\{\mathbb{E}_{\mathbb{P}_{\mathbf{s},\mathbf{r}}}[\log \sigma (T(\mathbf{s},\mathbf{r}))]+\mathbb{E}_{\mathbb{P}_{\mathbf{s}}\otimes\mathbb{P}_{\mathbf{r}}}[\log(1-\sigma(T(\mathbf{s},\mathbf{r})))]\right\}
    \label{eq:MINE_gan}
\end{align}
The inequality results from approximating $T(\cdot)$ with limited representation ability, which is a sub-class of projection $\mathbb{R}^c\times\mathbb{R}^c\to\mathbb{R}$. Our objective is to optimize $T(\cdot)$ to maximize the estimation bound in (\ref{eq:MINE_origin}), where $T(\cdot)$ is implemented by a learnable estimator $\Phi_{\mathrm{MI}}(\cdot)$, with the loss defined in~(\ref{eq:MINE_target}):
\begin{align}
    \mathcal{L}_{MI}=-\frac{1}{|\mathcal{P}_{\mathbf{s},\mathbf{r}}|}\sum_{(\mathbf{s},\mathbf{r})\in\mathcal{P}_{\mathbf{s},\mathbf{r}}}\log\sigma\left(\Phi_{\mathrm{MI}}(\mathbf{s},\mathbf{r})\right)-\frac{1}{|\mathcal{P}_{\mathbf{s}}| |\mathcal{P}_{\mathbf{r}}|}\sum_{\mathbf{s}\in\mathcal{P}_{\mathbf{s}},\mathbf{r}\in\mathcal{P}_{\mathbf{r}}}\log\left(1-\sigma\left(\Phi_{\mathrm{MI}}(\mathbf{s},\mathbf{r})\right)\right)
    \label{eq:MINE_target}
\end{align}
where $\mathcal{P}_{\mathbf{s},\mathbf{r}}$ denotes the set sampled from $\mathbb{P}_{\mathbf{s},\mathbf{r}}$ where $\mathbf{s},\mathbf{r}$ are visual features from two agents at the same location. $\mathcal{P}_{\mathbf{s}},\mathcal{P}_{\mathbf{r}}$ are sampled from the marginals where $\mathbf{s},\mathbf{r}$ are randomly combined.
We see that the mutual information estimator $\Phi_{\mathrm{MI}}(\cdot)$ actually serves as a discriminator. It predicts whether a feature pair $(\mathbf{s},\mathbf{r})$ is drawn from the joint distribution $\mathbb{P}_{\mathbf{s},\mathbf{r}}$, i.e., two agents’ observations at the same location that exhibit similar patterns, or from the product of marginals $\mathbb{P}_\mathbf{s}\otimes\mathbb{P}_\mathbf{r}$, in which observations are randomly paired and likely to differ in pattern due to non-corresponding locations.


\vspace{-1mm}
\textbf{Redundancy-less feature selection.} To reduce redundancy in transmission, we prioritize the message $\mathbf{s}$ with a low mutual information score $\Phi_{\mathrm{MI}}(\mathbf{s},\mathbf{r})$.
Specifically, we obtain the redundancy map $R_{s\to r}=\Phi_{\mathrm{MI}}(\hat{F}^q_{sc},F_r)\in \mathbb{R}^{h\times w}$ and derive the redundancy-less selection mask $M_{MI}=\mathbf{1}[R_{s\to r}<\tau_{MI}]$, where $\tau_{MI}$ is a redundancy threshold. The feature finally sent to the receiver is $Z_{s\to r}=M_{MI} \odot \hat{F}^q_{sc}$. This selection filters out the messages in $Z_{s\to r}$ that are already covered by $F_r$. 
The total communication volume of RDcomm is computed as $|D_s\odot M_c\odot M_{MI}|+|D_s^{base}\odot M_c|$. The first term measures the volume of the selected lossless information $Z_{s\to r}$, and the second term measures the cost for identifying redundancy and is much smaller than the first term by setting $\mathbf{B}_{base}$ with small codebook size $n$ and dimension $d$ with limited segments.

\vspace{-1mm}
\textbf{Message smoothing and fusion.} Note that $Z_{s\to r}$ is obtained under sparse masks $M_C$ and $M_{MI}$. Although it preserves salient information, the sparsity may degrade semantic content. We mitigate this by applying a UNet~\cite{ronneberger2015unet} $\Phi_{\mathrm{smth}}(\cdot)$ to smooth and dilate $Z_{s\to r}$, propagating sparse signals to neighboring regions. The receiver then obtains the enhanced perception results as $\bar{Y_r}=\Phi_{\mathrm{task}}(\Phi_{\mathrm{fusion}}(F_r,\Phi_{\mathrm{smth}}(Z_{s\to r)}))$, where $\Phi_{\mathrm{fusion}}(\cdot)$ is instantiated using the effective max-fusion operation following~\cite{hu2024codefilling}. 


\vspace{-2mm}
\subsection{Training}
\label{sec:training}
\vspace{-2mm}
We train RDcomm in three stages. First, we train the BEV encoder $\Phi_{\mathrm{bev}}$ and the task decoder $\Phi_{\mathrm{task}}$ with task loss $\mathcal{L}_{task}$, which corresponds to the CenterPoint loss for 3D detection and the per-pixel cross-entropy loss for BEV segmentation. Second, we train the vector quantization module $\Phi_{\mathrm{vq}}$ with both task loss $\mathcal{L}_{task}$ and feature reconstruction loss $\mathcal{L}_{recon}=\|F_s^q-F_s\|_2^2$. After that, the confidence frequency $p_c(\cdot)$ is updated. Finally, we train the mutual-information estimator $\Phi_{\mathrm{MI}}$ using $\mathcal{L}_{MI}$. In the later stages of training, the thresholds $\tau_{c},\tau_{MI}$ are randomly varied to facilitate bandwidth adaptation.

\vspace{-2mm}
\section{Experiments}
\vspace{-3mm}
To evaluate RDcomm, we conduct experiments on two representative collaborative perception tasks: 3D object detection and BEV semantic segmentation. To validate the effectiveness and generality of the proposed RDcomm, our evaluation spans both LiDAR and camera modalities, up to 5 collaborating agents, and varying bandwidth constraints.



\vspace{-2mm}
\textbf{Collaborative 3D detection.} We evaluate collaborative 3D detection on two datasets: the real-world DAIR-V2X~\cite{YuDAIRV2X:CVPR22} and the simulated OPV2V~\cite{XuOPV2V:ICRA22}. DAIR-V2X is a vehicle-to-infrastructure dataset with 9K frames of 2-agent collaboration between a vehicle and a roadside unit (RSU). Each agent is equipped with a LiDAR and a 1920×1080 camera, where the RSU uses a 300-channel LiDAR and the vehicle a 40-channel LiDAR. OPV2V is a vehicle-to-vehicle dataset simulated with CARLA~\cite{dosovitskiy2017carla}, containing 12K frames. Our experiments involve up to 3 agents, each equipped with a 64-channel LiDAR and four RGB cameras at 800×600 resolution. We evaluate both LiDAR and camera modalities and report Average Precision (AP) at IoU thresholds of 30\%, 50\%, and 70\%. Following~\cite{lu2024HEAL}, the perception range is set to 204.8m×102.4m.

\vspace{-2mm}
\textbf{Collaborative BEV segmentation.} We evaluate BEV semantic segmentation on OPV2V dataset following CoBEVT~\cite{XuCoBEVT:CoRL22}. Each agent predicts a BEV semantic occupancy map with camera inputs, ground truth classes include dynamic vehicles, drivable area, and lane. We involve collaboration among up to 5 agents. Performance is measured by Intersection-over-Union (IoU) between predictions and ground-truth BEV labels, with the perception range fixed at 100m×100m.

\vspace{-3mm}
\subsection{Quantitative analysis}
\vspace{-2mm}

\begin{figure}[!t]
  \centering
  \begin{subfigure}{0.24\linewidth}
    \includegraphics[width=1.0\linewidth]{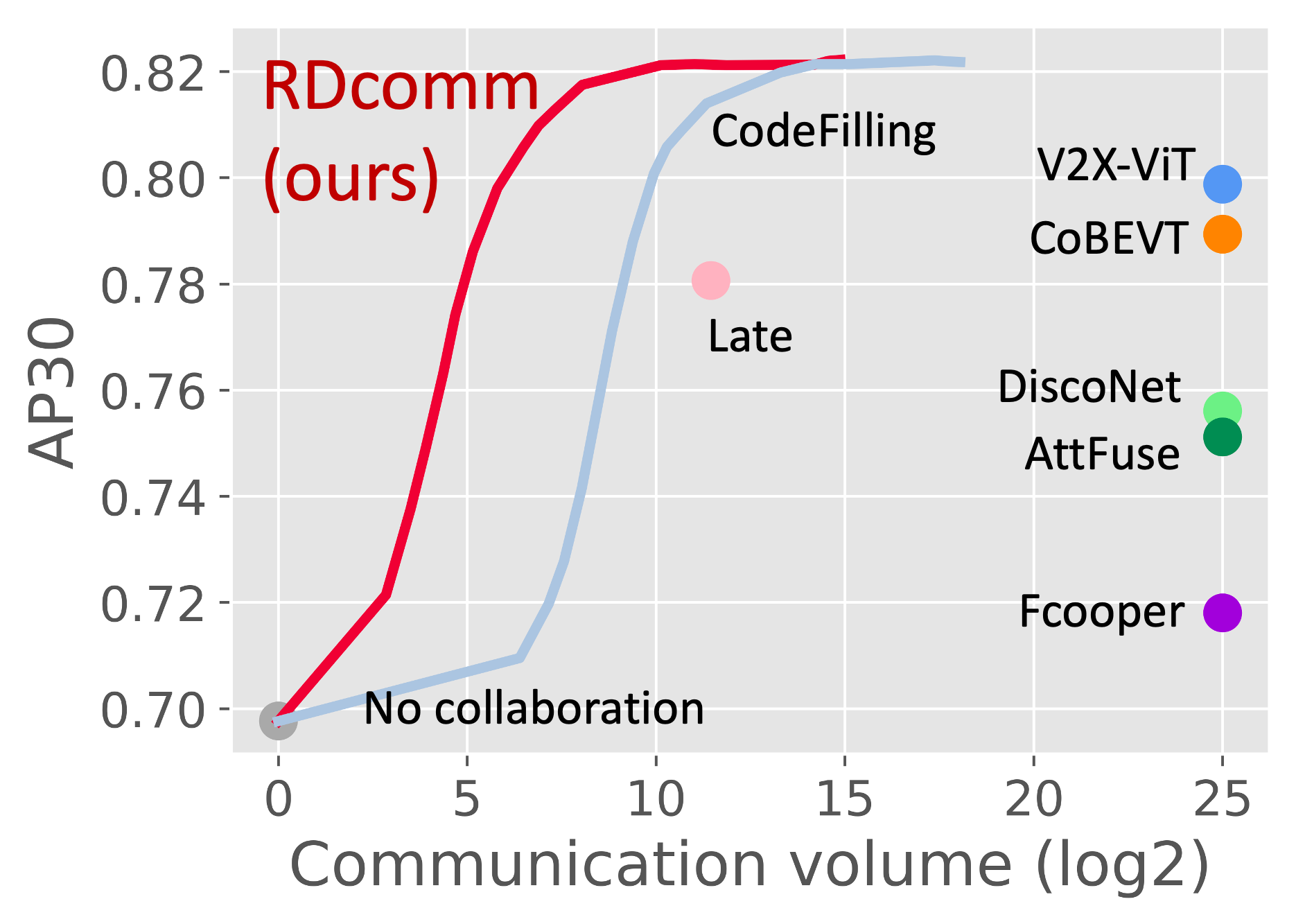}
    \vspace{-6mm}
    \caption{DAIR-V2X LiDAR}
  \end{subfigure}
  \begin{subfigure}{0.24\linewidth}
    \raisebox{0.00cm}{\includegraphics[width=1.0\linewidth]{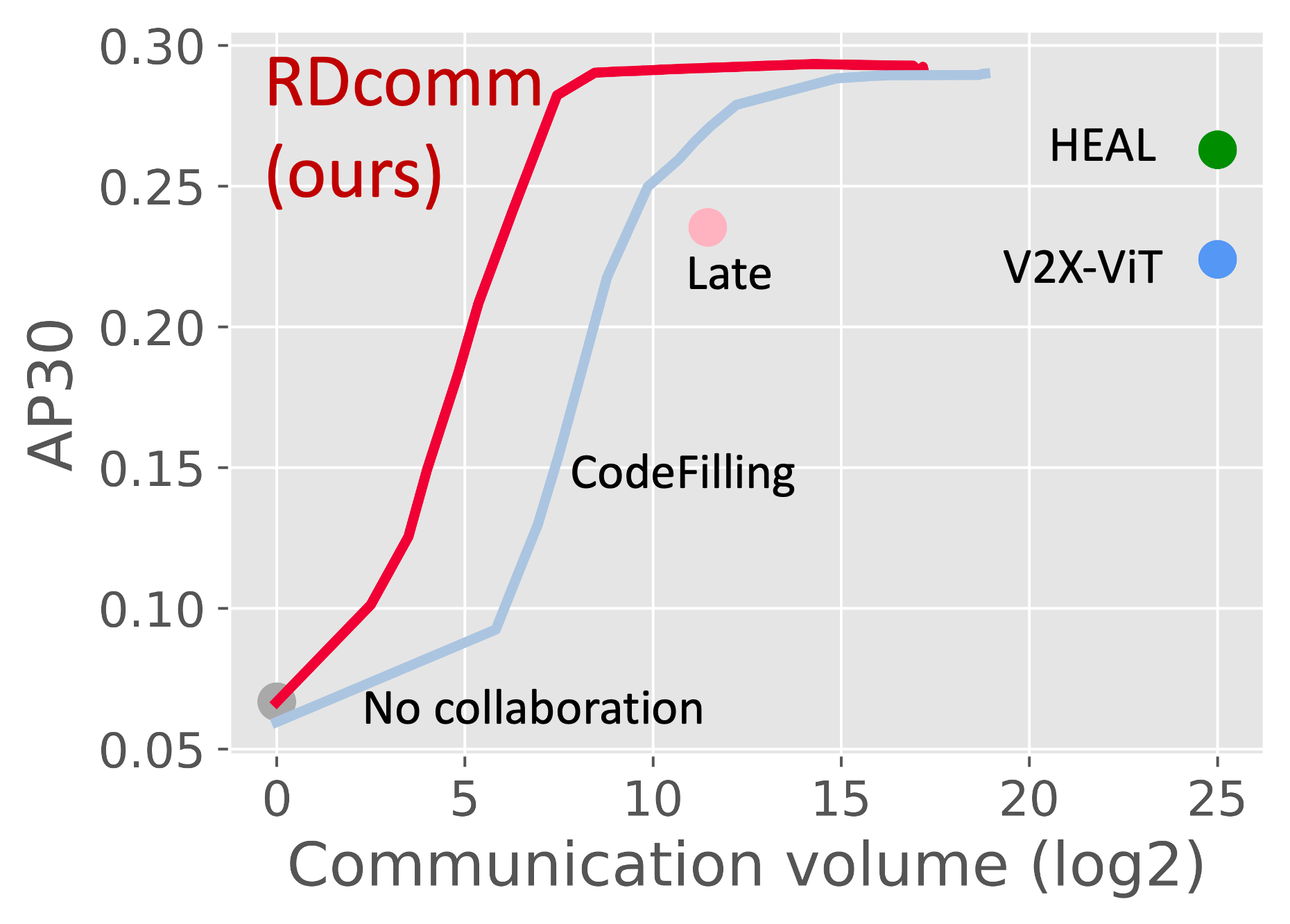}}
    \vspace{-6mm}
    \caption{DAIR-V2X camera}
  \end{subfigure}
  \begin{subfigure}{0.24\linewidth}
    \raisebox{0.00cm}{\includegraphics[width=1.0\linewidth]{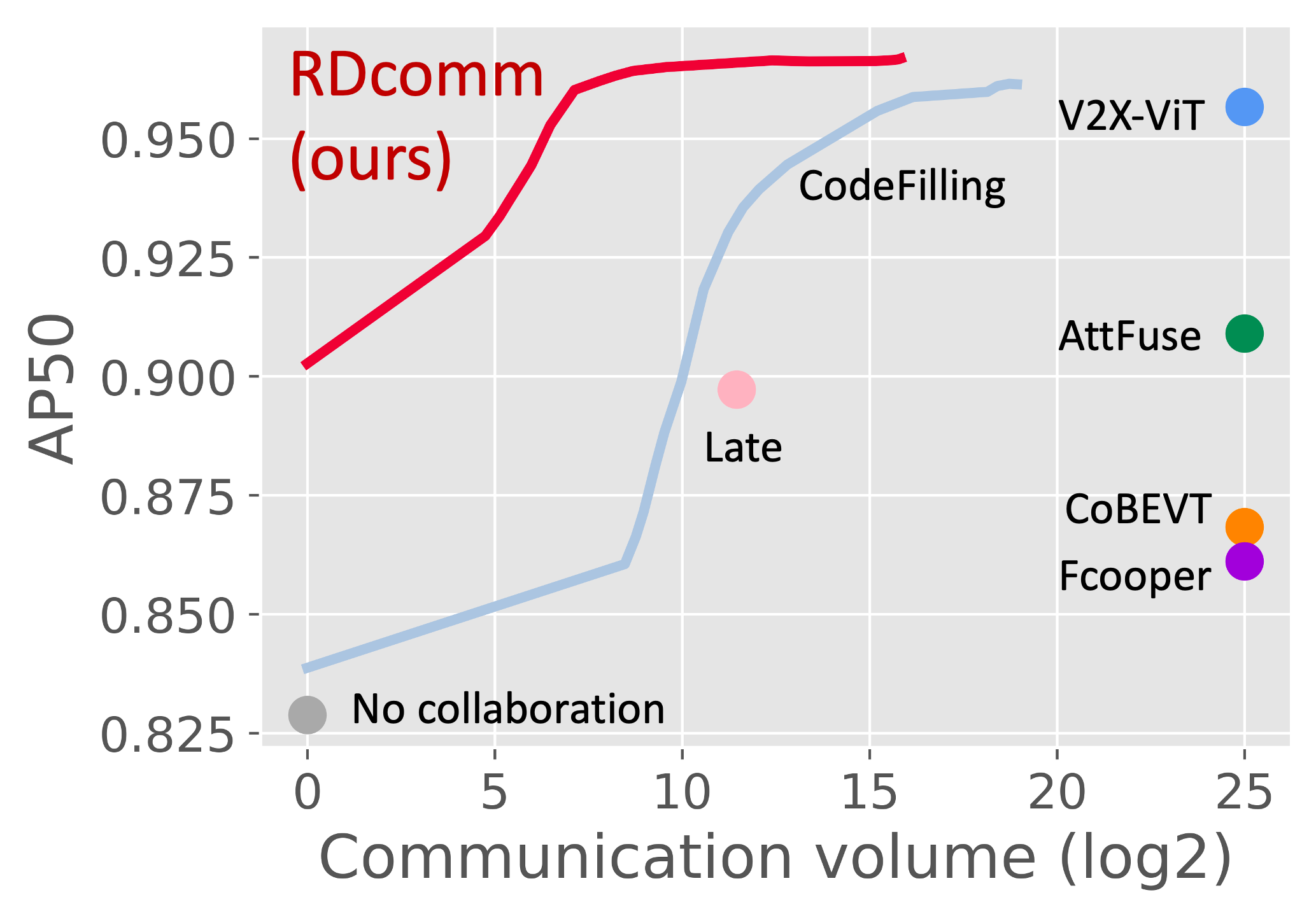}}
    \vspace{-6mm}
    \caption{OPV2V LiDAR}
  \end{subfigure}
  \begin{subfigure}{0.24\linewidth}
    \raisebox{0.00cm}{\includegraphics[width=1.0\linewidth]{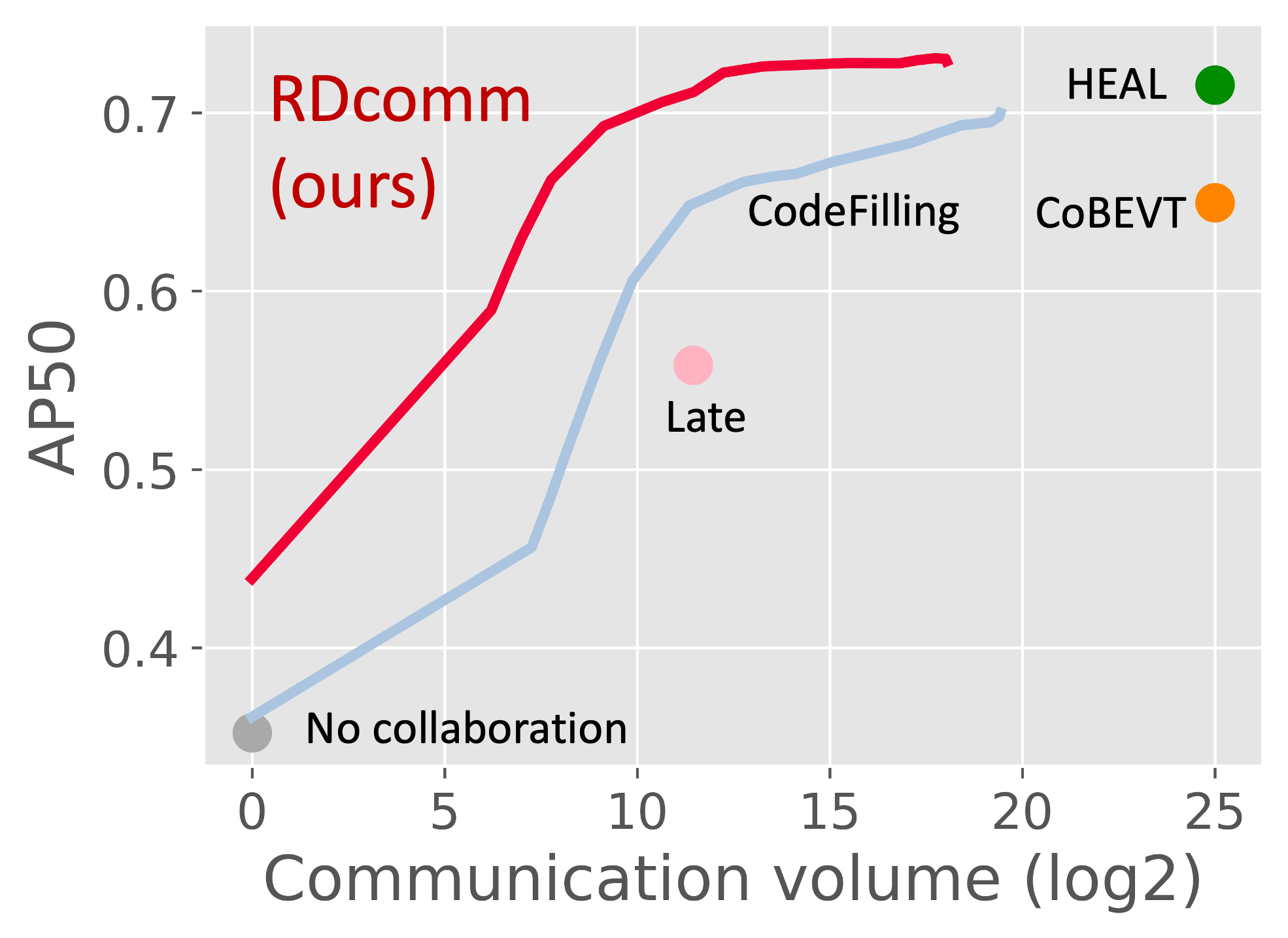}}
    \vspace{-6mm}
    \caption{OPV2V camera}
  \end{subfigure}
  \vspace{-3mm}
  \caption{\small RDcomm achieves the best performance–communication trade-off in 3D detection, across DAIR-V2X/OPV2V datasets with LiDAR/camera input modalities.}
  \vspace{-3mm}
  \label{fig:det_tradeoff}
\end{figure}

\begin{figure}[!t]
  \centering
  \begin{subfigure}{0.24\linewidth}
    \includegraphics[width=1.0\linewidth]{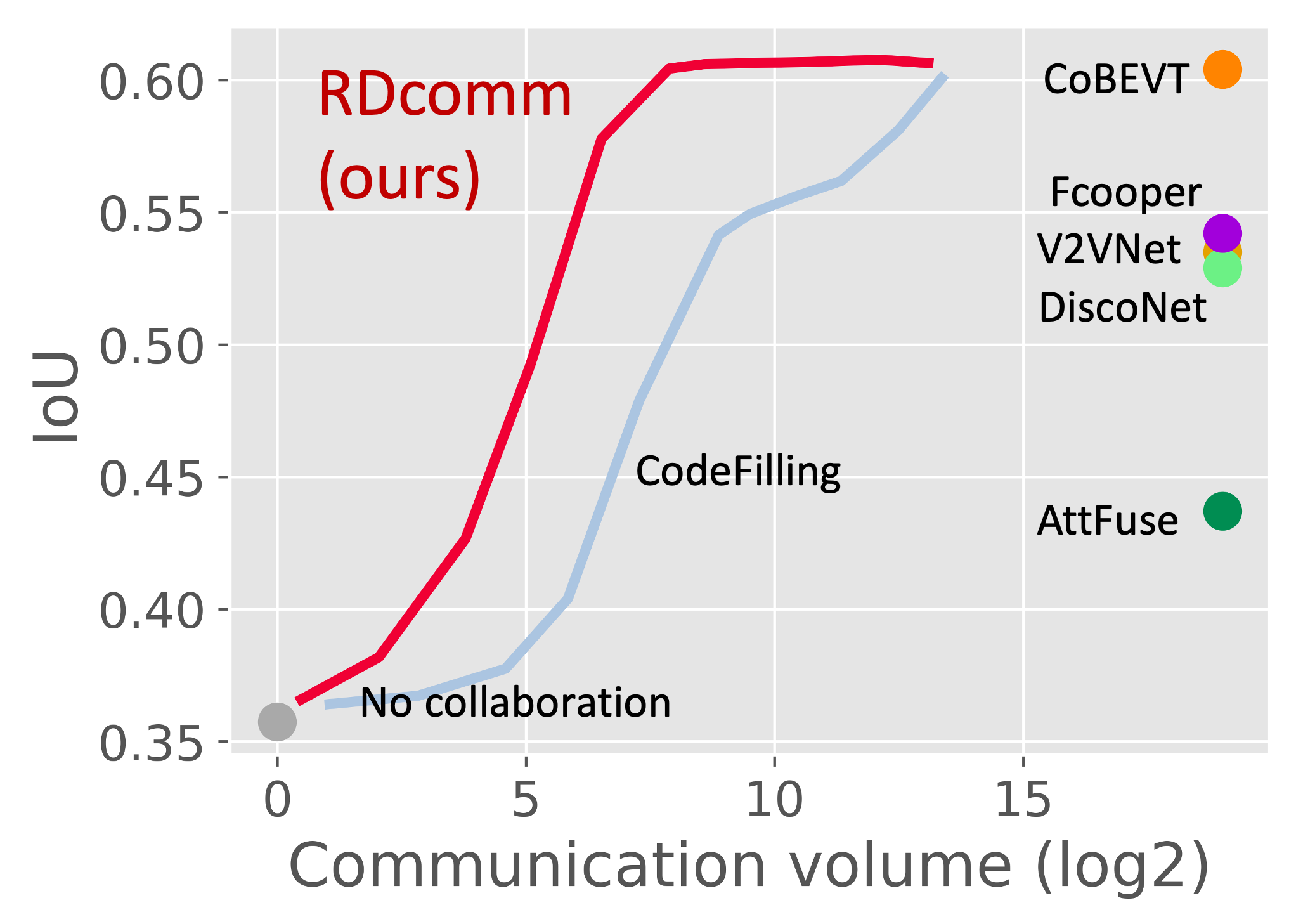}
    \vspace{-6mm}
    \caption{Vehicle}
  \end{subfigure}
  \begin{subfigure}{0.24\linewidth}
    \raisebox{0.00cm}{\includegraphics[width=1.0\linewidth]{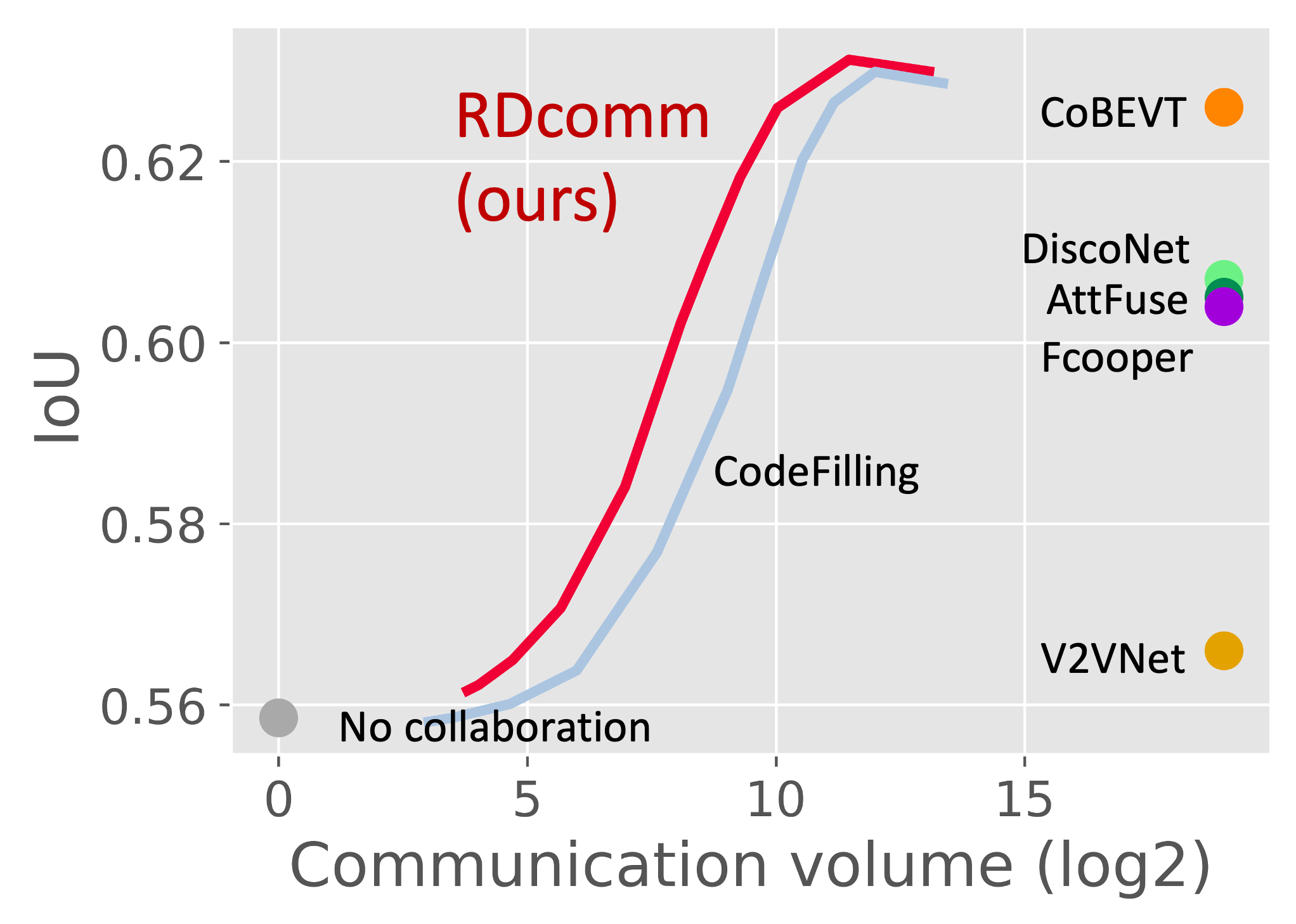}}
    \vspace{-6mm}
    \caption{Drive area}
  \end{subfigure}
  \begin{subfigure}{0.24\linewidth}
    \raisebox{0.00cm}{\includegraphics[width=1.0\linewidth]{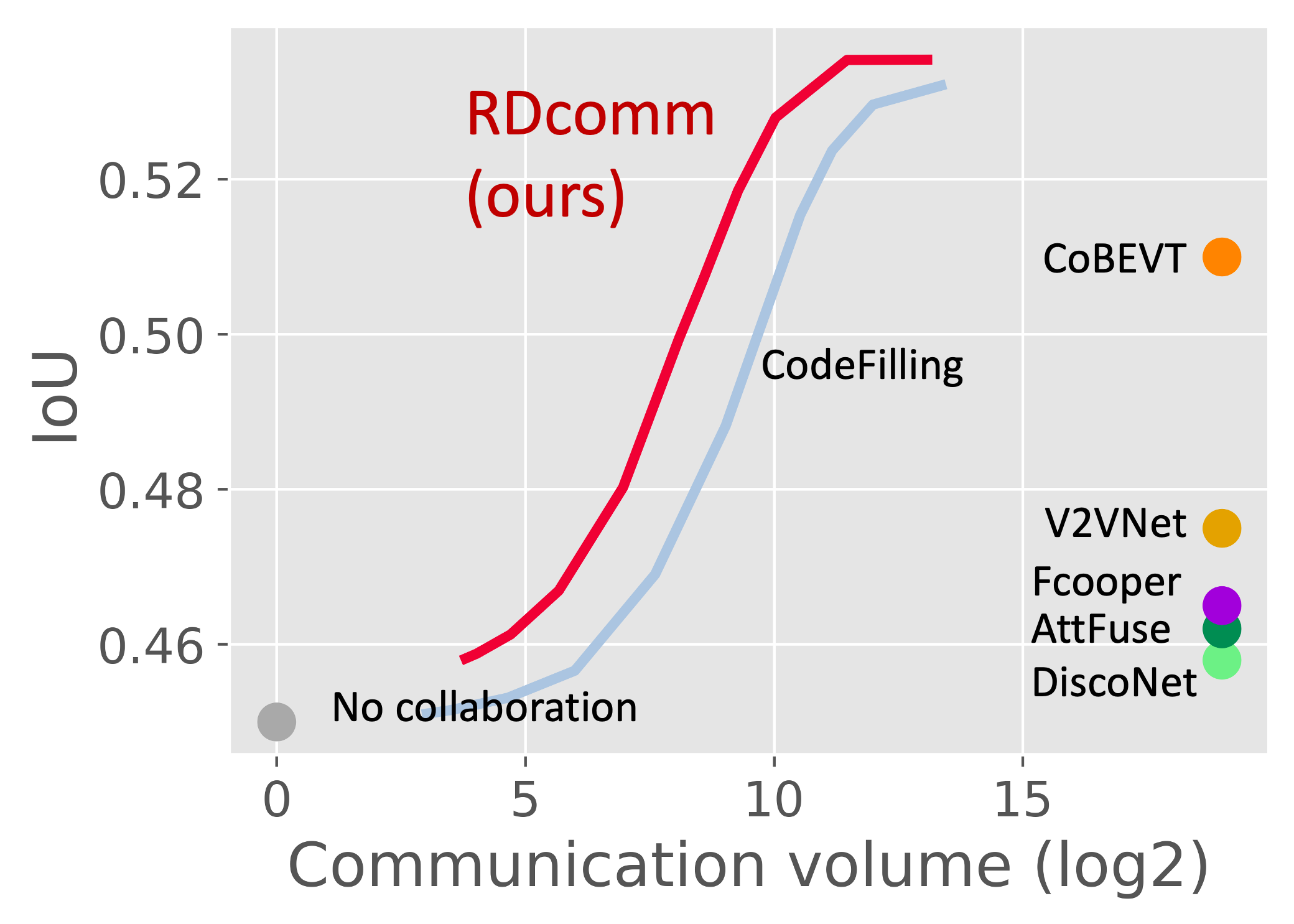}}
    \vspace{-6mm}
    \caption{Lane}
  \end{subfigure}
  \begin{subfigure}{0.24\linewidth}
    \raisebox{0.00cm}{\includegraphics[width=1.0\linewidth]{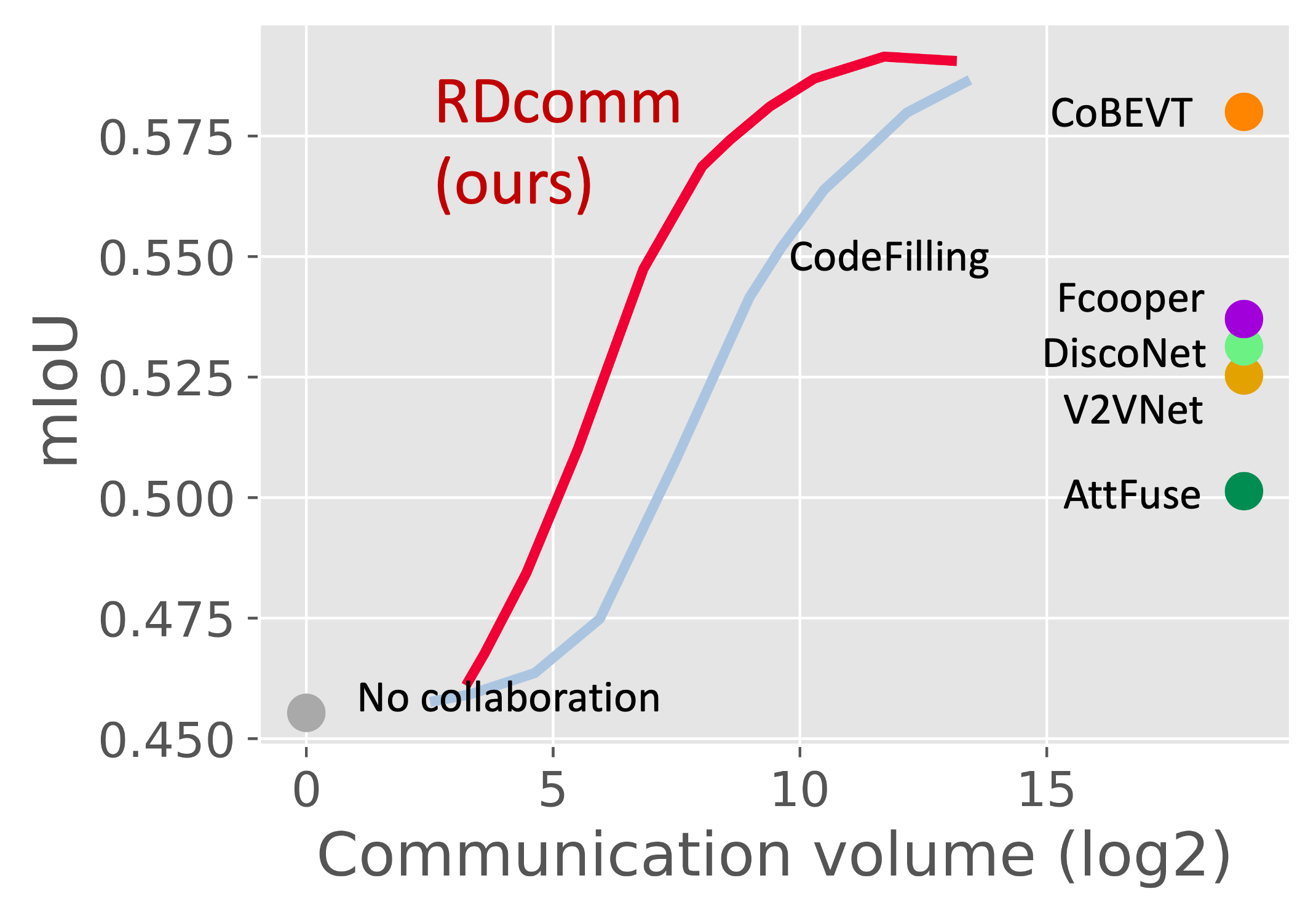}}
    \vspace{-6mm}
    \caption{Mean}
  \end{subfigure}
  \vspace{-3mm}
  \caption{\small RDcomm achieves the best performance–communication trade-off in BEV segmentation.}
  \vspace{-6mm}
  \label{fig:seg_tradeoff}
\end{figure}

\textbf{Benchmark comparison.} Fig. \ref{fig:det_tradeoff} and Fig. \ref{fig:seg_tradeoff} compare RDcomm with previous collaborative perception methods in terms of the trade-off between perception performance and communication volume on 3D detection and BEV segmentation. The baselines include no collaboration, CodeFilling~\cite{hu2024codefilling}, V2X-ViT~\cite{xu2022v2xvit}, CoBEVT~\cite{XuCoBEVT:CoRL22}, DiscoNet~\cite{li2021disconet}, AttFuse~\cite{XuOPV2V:ICRA22}, V2VNet~\cite{wang2020v2vnet}, Fcooper~\cite{chen2019fcooper}, HEAL~\cite{lu2024HEAL}, and late Fusion (directly share the final perception results). We see that: i) RDcomm consistently delivers a superior perception–communication trade-off across all bandwidth settings for both detection and segmentation; the gains persist across LiDAR/camera modalities and multiple semantic classes.
ii) Under extreme bandwidth constraints, RDcomm achieves larger gains than prior methods: for detection, +11.49/+19.82\% (LiDAR/camera) on DAIR-V2X and +12.01/+22.92\% on OPV2V with a 50K times reduction relative to uncompressed features; for segmentation, +5.69\% mIoU at a 1K times reduction.
iii) RDcomm outperforms previous communication-efficient SOTA, CodeFilling, with significantly reduced communication cost in detection: 15/13 times less (LiDAR/camera) on DAIR-V2X and 54/108 times less on OPV2V; and 8 times less for segmentation.

\begin{figure}[!h]
  \centering
  \vspace{-2mm}
  \begin{subfigure}{0.48\linewidth} 
    \centering
    \begin{subfigure}{0.5\linewidth}
      \includegraphics[width=\linewidth]{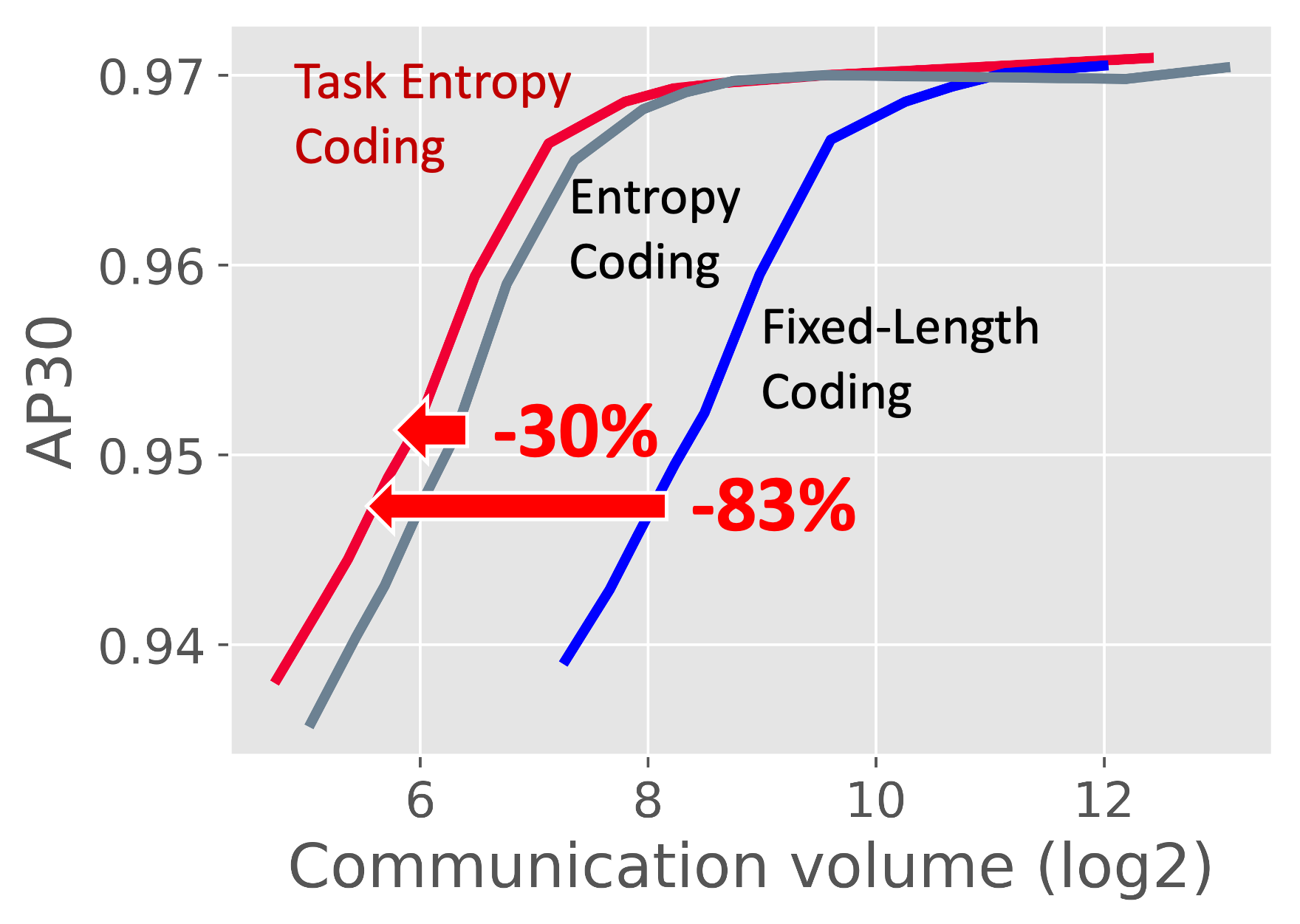}
    \end{subfigure}%
    \begin{subfigure}{0.5\linewidth}
      \includegraphics[width=\linewidth]{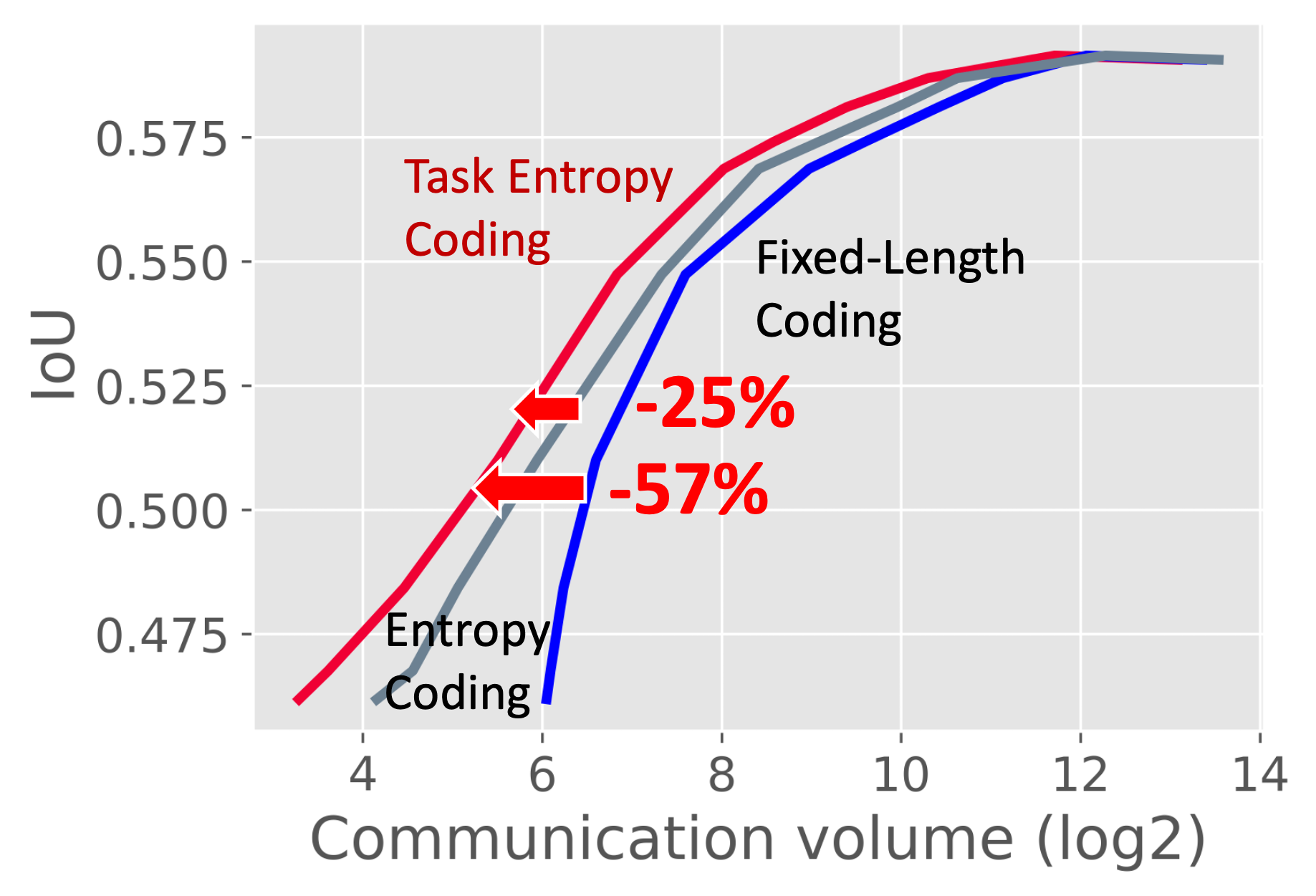}
    \end{subfigure}
    \vspace{-6mm}
    \caption{Ablation on the coding method} 
    \label{fig:ablation_coding}
  \end{subfigure}%
  \begin{subfigure}{0.48\linewidth} 
    \centering
    \begin{subfigure}{0.5\linewidth}
      \includegraphics[width=\linewidth]{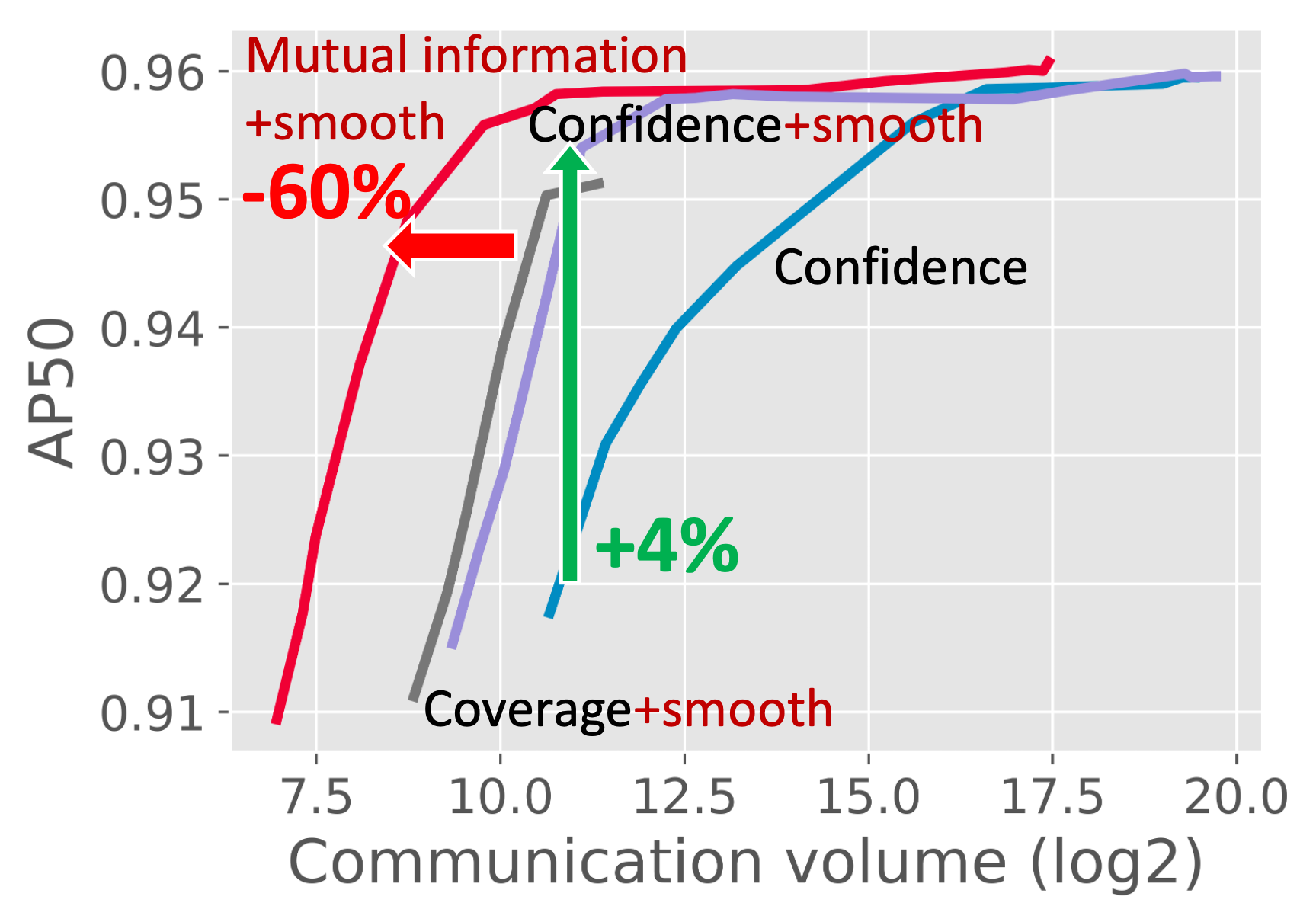}
    \end{subfigure}%
    \begin{subfigure}{0.5\linewidth}
      \includegraphics[width=\linewidth]{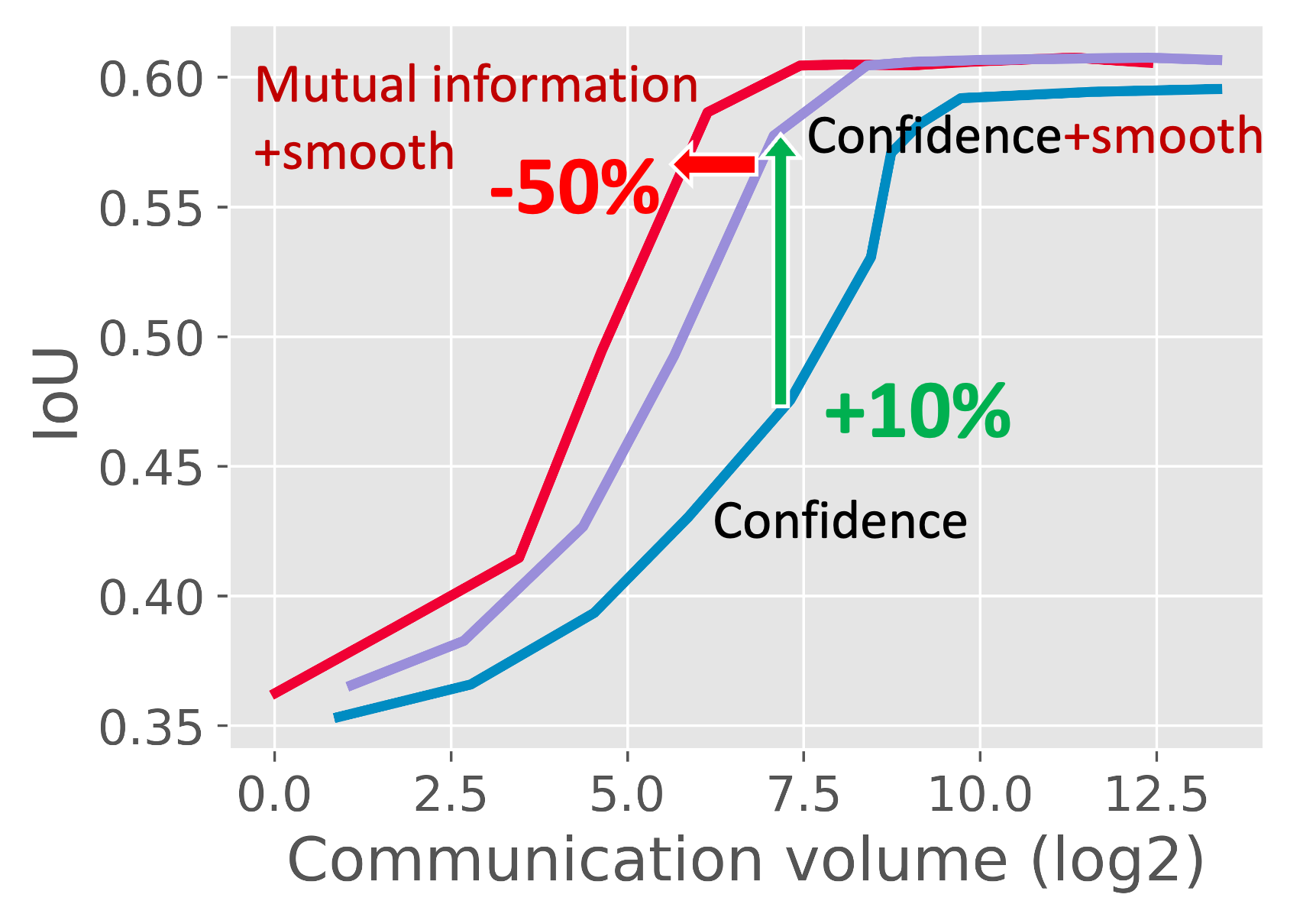}
    \end{subfigure}
    \vspace{-6mm}
    \caption{Ablation on the selection method} 
    \label{fig:ablation_selection}
  \end{subfigure}
  \vspace{-4mm}
  \caption{\small Ablation study on OPV2V detection/segmentation, evaluating coding and selection methods.}
  \label{fig:ablation}
  \vspace{-4mm}
\end{figure}

\vspace{-2mm}
\textbf{Ablation on coding method.} Fig.~\ref{fig:ablation_coding} compares the proposed task entropy coding in RDcomm against: i) classic entropy coding weighted by occurrence frequency~\cite{huffman2007method}; ii) fixed-length coding~\cite{hu2024codefilling}, where code length is $\log2$ of the codebook volume. We see that task entropy coding saves 83/57\%(detection/segmentation) in communication volume compared to fixed-length coding, and 30/25\% compared to occurrence-driven entropy coding. The reason is that our task entropy coding prioritizes pragmatic-rich codes by assigning them shorter codewords, whereas classic entropy coding may waste short codewords on high-frequency but pragmatically weak codes. Theoretically, we use confidence mask $M_C$ to reduce the entropy $\mathrm{H}(Z_{s\to r}|Y)$, and entropy coding weighted by confidence frequency $p_c(\cdot)$ shortens the expected code length toward the entropy limit implied by this entropy.


\vspace{-2mm}
\textbf{Ablation on selection method.} Fig.~\ref{fig:ablation_selection} compares our mutual-information–driven (MI) redundancy selection against other redundancy selection: i) confidence-based~\cite{hu2024codefilling}; ii) LiDAR-coverage–based~\cite{xu2025cosdh}. 
Our MI selection reduces communication volume by 60/50\% (detection/segmentation) relative to these baselines. The gain arises since MI leverages a pragmatic yet lightweight abstraction to identify redundancy, providing richer cues than one-dimensional confidence or coverage signals. Theoretically, the mutual information estimator encourages agents to transmit $Z_{s\to r}$ with low $\mathrm{I}(Z_{s\to r},F_r)$ relative to the receiver’s feature $F_r$. Meanwhile, we observe that the smoothing module improves AP50 by 4\% and IoU by 10\%, demonstrating its effectiveness in mitigating sparsity under high selection rates.

\begin{wraptable}[8]{r}{0.38\linewidth}
\centering
\caption{\small Segmentation performance with compression. bpp: bits-per-pixel.}
\vspace{-2mm}
\label{tab:lossless_bits}
\resizebox{\linewidth}{!}{
\begin{tabular}{lll}
\toprule
Method & mIoU$\uparrow$ & bpp$\downarrow$ \\ \midrule
VQVAE-32  & 0.38 \textcolor{gray}{(82\%)} & 10 \\
VQVAE-128 & 0.40 \textcolor{gray}{(87\%)} & 14 \\
RDcomm-128  & 0.44 \textcolor{gray}{(95\%)} & 4 \\ \hline
No compression  & 0.46 \textcolor{gray}{(100\%)} & 4096 \\ \bottomrule
\end{tabular}
}
\end{wraptable}

\textbf{Lossless bit-rate.} We compare the bit-rate of RDcomm with the optimal bit-rate $\mathrm{Rate}(\delta)$ in ~(\ref{eq:rate_distortion_collab_final_0}) under lossless pragmatic compression, i.e., $\delta=0$. To reduce the error in estimating $\mathrm{Rate}(0)$, we exclude the receiver’s information $X_r$ in the experiment; the optimal rate is then $\mathrm{I}(Y;X_s)$, which is tightly upper-bounded by $\mathrm{H}(Y)\approx\log_2(4)$, i.e., 2 bits-per-pixel (bpp). Here we define "lossless" as a performance drop of less than $5\%$. Tab.~\ref{tab:lossless_bits} compares RDcomm’s BEV segmentation performance with the no-compression scheme. RDcomm attains 95\% of mIoU with 4 bpp, close to the 2 bpp upper bound. Here bpp describes BEV feature. We also report VQVAE~\cite{van2017vqvae} results (codebook sizes 32/128, segment number 2), where RDcomm uses a codebook size of 128. RDcomm delivers higher accuracy at a lower rate, indicating that effective pragmatic compression cannot be achieved by merely tuning codebook size,  requires selective transmission and shorter codewords for task-relevant codes.



\begin{wraptable}{r}{0.40\linewidth}
\centering
\vspace{-3mm}
\caption{\small Allocation of communication volume.}
\vspace{-2mm}
\label{tab:two_stage}
\resizebox{\linewidth}{!}{
\begin{tabular}{llll}
\toprule
AP30 & 0.82 & 0.79 & 0.76 \\ \hline
total bits & 9054 & 449 & 166 \\
abstract bits & 882\textcolor{gray}{(9\%)} & 49\textcolor{gray}{(11\%)} & 19\textcolor{gray}{(11\%)} \\ \bottomrule
\end{tabular}
}
\vspace{-2mm}
\end{wraptable}
\textbf{Cost for transmitting abstract.} 
Tab.~\ref{tab:two_stage} reports the share of bandwidth consumed by transmitting the pragmatic abstraction $\hat{F}^q_{sc}$ on DAIR-V2X detection. 
We observe that abstraction transmission accounts for only 9\%–11\% of the total communication volume, yet is effective to identify redundancy as showed in Fig.~\ref{fig:ablation_selection}.





\begin{figure}[!h]
\centering
\includegraphics[width=0.97\linewidth]{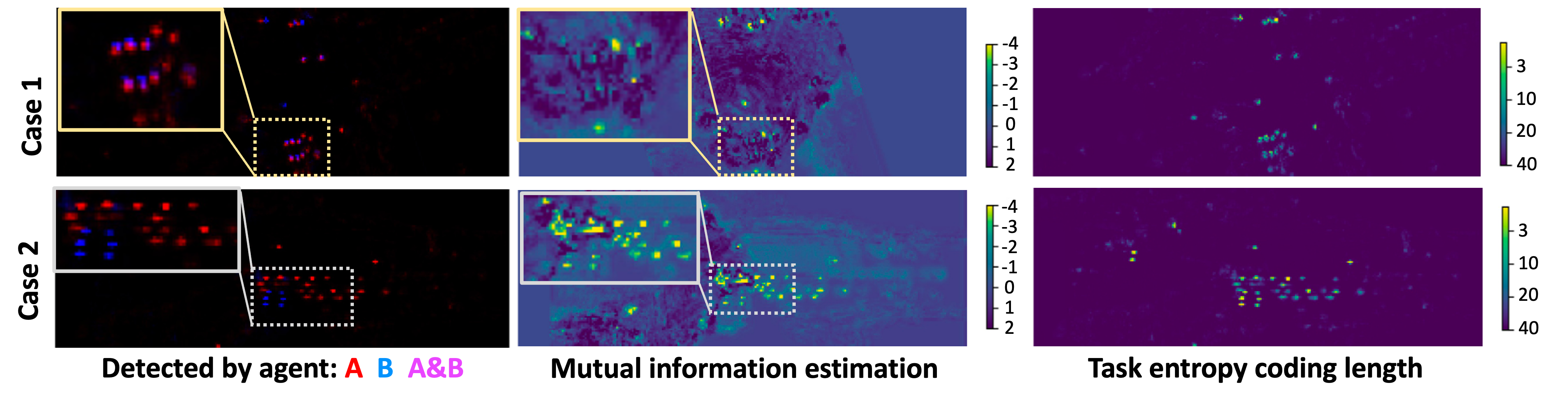}
\vspace{-4mm}
\caption{\small Visualization of mutual information estimation and task entropy coding length on DAIR-V2X.}
\label{fig:demo}
\vspace{-2mm}
\end{figure}

\subsection{Qualitative analysis}
\label{sec:qualitative}
Fig.~\ref{fig:demo} visualizes mutual-information estimates and task-entropy code lengths in two cases.
In case 1, both agents A and B detect the same vehicles, forming high–mutual-information regions.
In case 2, A and B detect different vehicles, yielding low–mutual-information regions that are prioritized for sharing.
We also observe that task-entropy coding assigns short codewords to task-relevant regions. Long codewords are assigned to background areas, which are omitted when bandwidth is limited.

\section{Conclusions}
This work investigates the trade-off between task performance and communication volume from an information-theoretic perspective. We formulate a pragmatic rate–distortion theory for collaborative perception, deriving the optimal bit rate for message transmission and two necessary conditions for optimal compression: pragmatic-relevant and redundancy-less. Guided by these two conditions, we propose RDcomm, a communication-efficient collaborative perception method with two novel components: i) task entropy discrete coding and ii) mutual-information-driven message selection. Experiments covering both detection and segmentation show that RDcomm achieves state-of-the-art perception–communication trade-offs across both LiDAR and camera modalities.

\textbf{Limitations and future work.} Our study focuses on perception tasks. Future directions include extending the framework to broader tasks such as navigation, manipulation, and scene captioning, and incorporating additional modalities such as motion and language.

\textbf{Reproducibility Statement.}
We have taken multiple steps to ensure the reproducibility of our work. The detailed formulations of our theoretical results, including all assumptions and proofs, are provided in Appendix~\ref{sec:proofs}. The design of RDcomm and its components is described in Sec.~\ref{sec:rdcomm}, with training details in Sec.~\ref{sec:training} and backbone details reported in Appendix.~\ref{sec:single-agent-perception}. The datasets used in our experiments (DAIR-V2X and OPV2V) are publicly available. We will release our source code and configuration files in the camera-ready version to further facilitate reproducibility.



\bibliography{iclr2026_conference}

\begin{thebibliography}{30}
\providecommand{\natexlab}[1]{#1}
\providecommand{\url}[1]{\texttt{#1}}
\expandafter\ifx\csname urlstyle\endcsname\relax
  \providecommand{\doi}[1]{doi: #1}\else
  \providecommand{\doi}{doi: \begingroup \urlstyle{rm}\Url}\fi

\bibitem[Ball{\'e} et~al.(2018)Ball{\'e}, Minnen, Singh, Hwang, and Johnston]{balle2018variational}
Johannes Ball{\'e}, David Minnen, Saurabh Singh, Sung~Jin Hwang, and Nick Johnston.
\newblock Variational image compression with a scale hyperprior.
\newblock \emph{arXiv preprint arXiv:1802.01436}, 2018.

\bibitem[Belghazi et~al.(2018)Belghazi, Baratin, Rajeshwar, Ozair, Bengio, Courville, and Hjelm]{belghazi2018MINE}
Mohamed~Ishmael Belghazi, Aristide Baratin, Sai Rajeshwar, Sherjil Ozair, Yoshua Bengio, Aaron Courville, and Devon Hjelm.
\newblock Mutual information neural estimation.
\newblock In \emph{International conference on machine learning}, pp.\  531--540. PMLR, 2018.

\bibitem[Blau \& Michaeli(2019)Blau and Michaeli]{blau2019rethinking_distortion}
Yochai Blau and Tomer Michaeli.
\newblock Rethinking lossy compression: The rate-distortion-perception tradeoff.
\newblock In \emph{International Conference on Machine Learning}, pp.\  675--685. PMLR, 2019.

\bibitem[Chen et~al.(2019)Chen, Ma, Tang, Guo, Yang, and Fu]{chen2019fcooper}
Qi~Chen, Xu~Ma, Sihai Tang, Jingda Guo, Qing Yang, and Song Fu.
\newblock F-cooper: Feature based cooperative perception for autonomous vehicle edge computing system using 3d point clouds.
\newblock In \emph{Proceedings of the 4th ACM/IEEE Symposium on Edge Computing}, pp.\  88--100, 2019.

\bibitem[Cover(1999)]{cover1999elements_infor_theory}
Thomas~M Cover.
\newblock \emph{Elements of information theory}.
\newblock John Wiley \& Sons, 1999.

\bibitem[Dosovitskiy et~al.(2017)Dosovitskiy, Ros, Codevilla, Lopez, and Koltun]{dosovitskiy2017carla}
Alexey Dosovitskiy, German Ros, Felipe Codevilla, Antonio Lopez, and Vladlen Koltun.
\newblock Carla: An open urban driving simulator.
\newblock In \emph{Conference on robot learning}, pp.\  1--16. PMLR, 2017.

\bibitem[Dubois et~al.(2021)Dubois, Bloem-Reddy, Ullrich, and Maddison]{dubois2021lossy}
Yann Dubois, Benjamin Bloem-Reddy, Karen Ullrich, and Chris~J Maddison.
\newblock Lossy compression for lossless prediction.
\newblock \emph{Advances in Neural Information Processing Systems}, 34:\penalty0 14014--14028, 2021.

\bibitem[Han et~al.(2023)Han, Zhang, Li, Jin, Lang, and Li]{han2023survey}
Yushan Han, Hui Zhang, Huifang Li, Yi~Jin, Congyan Lang, and Yidong Li.
\newblock Collaborative perception in autonomous driving: Methods, datasets, and challenges.
\newblock \emph{IEEE Intelligent Transportation Systems Magazine}, 15\penalty0 (6):\penalty0 131--151, 2023.

\bibitem[Hu et~al.(2022)Hu, Fang, Lei, Zhong, and Chen]{hu2022where2comm}
Yue Hu, Shaoheng Fang, Zixing Lei, Yiqi Zhong, and Siheng Chen.
\newblock Where2comm: Communication-efficient collaborative perception via spatial confidence maps.
\newblock \emph{Advances in neural information processing systems}, 35:\penalty0 4874--4886, 2022.

\bibitem[Hu et~al.(2024)Hu, Peng, Liu, Ge, Liu, and Chen]{hu2024codefilling}
Yue Hu, Juntong Peng, Sifei Liu, Junhao Ge, Si~Liu, and Siheng Chen.
\newblock Communication-efficient collaborative perception via information filling with codebook.
\newblock In \emph{Proceedings of the IEEE/CVF Conference on Computer Vision and Pattern Recognition}, pp.\  15481--15490, 2024.

\bibitem[Huffman(2007)]{huffman2007method}
David~A Huffman.
\newblock A method for the construction of minimum-redundancy codes.
\newblock \emph{Proceedings of the IRE}, 40\penalty0 (9):\penalty0 1098--1101, 2007.

\bibitem[Lang et~al.(2019)Lang, Vora, Caesar, Zhou, Yang, and Beijbom]{lang2019pointpillars}
Alex~H Lang, Sourabh Vora, Holger Caesar, Lubing Zhou, Jiong Yang, and Oscar Beijbom.
\newblock Pointpillars: Fast encoders for object detection from point clouds.
\newblock In \emph{Proceedings of the IEEE/CVF conference on computer vision and pattern recognition}, pp.\  12697--12705, 2019.

\bibitem[Li et~al.(2020)Li, Chen, Zhang, and Tsang]{li2020gxn}
Maosen Li, Siheng Chen, Ya~Zhang, and Ivor Tsang.
\newblock Graph cross networks with vertex infomax pooling.
\newblock \emph{Advances in neural information processing systems}, 33:\penalty0 14093--14105, 2020.

\bibitem[Li et~al.(2021)Li, Ren, Wu, Chen, Feng, and Zhang]{li2021disconet}
Yiming Li, Shunli Ren, Pengxiang Wu, Siheng Chen, Chen Feng, and Wenjun Zhang.
\newblock Learning distilled collaboration graph for multi-agent perception.
\newblock \emph{Advances in Neural Information Processing Systems}, 34:\penalty0 29541--29552, 2021.

\bibitem[Lu et~al.(2023)Lu, Li, Liu, Dianati, Feng, Chen, and Wang]{lu2023robust}
Yifan Lu, Quanhao Li, Baoan Liu, Mehrdad Dianati, Chen Feng, Siheng Chen, and Yanfeng Wang.
\newblock Robust collaborative 3d object detection in presence of pose errors.
\newblock In \emph{2023 IEEE International Conference on Robotics and Automation (ICRA)}, pp.\  4812--4818. IEEE, 2023.

\bibitem[Lu et~al.(2024)Lu, Hu, Zhong, Wang, Wang, and Chen]{lu2024HEAL}
Yifan Lu, Yue Hu, Yiqi Zhong, Dequan Wang, Yanfeng Wang, and Siheng Chen.
\newblock An extensible framework for open heterogeneous collaborative perception.
\newblock \emph{arXiv preprint arXiv:2401.13964}, 2024.

\bibitem[Nowozin et~al.(2016)Nowozin, Cseke, and Tomioka]{nowozin2016fgan}
Sebastian Nowozin, Botond Cseke, and Ryota Tomioka.
\newblock f-gan: Training generative neural samplers using variational divergence minimization.
\newblock \emph{Advances in neural information processing systems}, 29, 2016.

\bibitem[Philion \& Fidler(2020)Philion and Fidler]{philion2020lss}
Jonah Philion and Sanja Fidler.
\newblock Lift, splat, shoot: Encoding images from arbitrary camera rigs by implicitly unprojecting to 3d.
\newblock In \emph{Computer Vision--ECCV 2020: 16th European Conference, Glasgow, UK, August 23--28, 2020, Proceedings, Part XIV 16}, pp.\  194--210. Springer, 2020.

\bibitem[Ronneberger et~al.(2015)Ronneberger, Fischer, and Brox]{ronneberger2015unet}
Olaf Ronneberger, Philipp Fischer, and Thomas Brox.
\newblock U-net: Convolutional networks for biomedical image segmentation.
\newblock In \emph{International Conference on Medical image computing and computer-assisted intervention}, pp.\  234--241. Springer, 2015.

\bibitem[Shannon et~al.(1959)]{shannon1959coding}
Claude~E Shannon et~al.
\newblock Coding theorems for a discrete source with a fidelity criterion.
\newblock \emph{IRE Nat. Conv. Rec}, 4\penalty0 (142-163):\penalty0 1, 1959.

\bibitem[Shao et~al.(2024)Shao, Li, and Zhang]{shao2024task}
Jiawei Shao, Teng Li, and Jun Zhang.
\newblock Task-oriented communication for vehicle-to-infrastructure cooperative perception.
\newblock In \emph{2024 IEEE 34th International Workshop on Machine Learning for Signal Processing (MLSP)}, pp.\  1--6. IEEE, 2024.

\bibitem[Van Den~Oord et~al.(2017)Van Den~Oord, Vinyals, et~al.]{van2017vqvae}
Aaron Van Den~Oord, Oriol Vinyals, et~al.
\newblock Neural discrete representation learning.
\newblock \emph{Advances in neural information processing systems}, 30, 2017.

\bibitem[Wang et~al.(2020)Wang, Manivasagam, Liang, Yang, Zeng, and Urtasun]{wang2020v2vnet}
Tsun-Hsuan Wang, Sivabalan Manivasagam, Ming Liang, Bin Yang, Wenyuan Zeng, and Raquel Urtasun.
\newblock V2vnet: Vehicle-to-vehicle communication for joint perception and prediction.
\newblock In \emph{Computer vision--ECCV 2020: 16th European conference, Glasgow, UK, August 23--28, 2020, proceedings, part II 16}, pp.\  605--621. Springer, 2020.

\bibitem[Xu et~al.(2025)Xu, Zhang, Cai, and Huang]{xu2025cosdh}
Junhao Xu, Yanan Zhang, Zhi Cai, and Di~Huang.
\newblock Cosdh: Communication-efficient collaborative perception via supply-demand awareness and intermediate-late hybridization.
\newblock In \emph{Proceedings of the Computer Vision and Pattern Recognition Conference}, pp.\  6834--6843, 2025.

\bibitem[Xu et~al.(2021)Xu, Xiang, Xia, Han, Liu, and Ma]{XuOPV2V:ICRA22}
Runsheng Xu, Hao Xiang, Xin Xia, Xu~Han, Jinlong Liu, and Jiaqi Ma.
\newblock Opv2v: An open benchmark dataset and fusion pipeline for perception with vehicle-to-vehicle communication.
\newblock \emph{2022 International Conference on Robotics and Automation (ICRA)}, pp.\  2583--2589, 2021.

\bibitem[Xu et~al.(2022{\natexlab{a}})Xu, Tu, Xiang, Shao, Zhou, and Ma]{XuCoBEVT:CoRL22}
Runsheng Xu, Zhengzhong Tu, Hao Xiang, Wei Shao, Bolei Zhou, and Jiaqi Ma.
\newblock {CoBEVT}: Cooperative bird's eye view semantic segmentation with sparse transformers.
\newblock \emph{CoRL}, 2022{\natexlab{a}}.

\bibitem[Xu et~al.(2022{\natexlab{b}})Xu, Xiang, Tu, Xia, Yang, and Ma]{xu2022v2xvit}
Runsheng Xu, Hao Xiang, Zhengzhong Tu, Xin Xia, Ming-Hsuan Yang, and Jiaqi Ma.
\newblock V2x-vit: Vehicle-to-everything cooperative perception with vision transformer.
\newblock In \emph{European conference on computer vision}, pp.\  107--124. Springer, 2022{\natexlab{b}}.

\bibitem[Yin et~al.(2021)Yin, Zhou, and Krahenbuhl]{yin2021center}
Tianwei Yin, Xingyi Zhou, and Philipp Krahenbuhl.
\newblock Center-based 3d object detection and tracking.
\newblock In \emph{Proceedings of the IEEE/CVF conference on computer vision and pattern recognition}, pp.\  11784--11793, 2021.

\bibitem[Yu et~al.(2022)Yu, Luo, Shu, Huo, Yang, Shi, Guo, Li, Hu, Yuan, et~al.]{YuDAIRV2X:CVPR22}
Haibao Yu, Yizhen Luo, Mao Shu, Yiyi Huo, Zebang Yang, Yifeng Shi, Zhenglong Guo, Hanyu Li, Xing Hu, Jirui Yuan, et~al.
\newblock {DAIR-V2X}: A large-scale dataset for vehicle-infrastructure cooperative 3d object detection.
\newblock \emph{In Proceedings of the IEEE/CVF Conference on computer vision and pattern recognition (CVPR)}, 2022.

\bibitem[Zhu et~al.(2022)Zhu, Song, Gao, Zheng, and Shen]{zhu2022unified}
Xiaosu Zhu, Jingkuan Song, Lianli Gao, Feng Zheng, and Heng~Tao Shen.
\newblock Unified multivariate gaussian mixture for efficient neural image compression.
\newblock In \emph{Proceedings of the IEEE/CVF Conference on Computer Vision and Pattern Recognition}, pp.\  17612--17621, 2022.

\end{thebibliography}
\bibliographystyle{iclr2026_conference}

\appendix

\section{Appendix}

\subsection{Statements}

\textbf{LLM Usage.}
We used an LLM (ChatGPT) solely for language refinement, such as improving grammar, clarity, and readability of sentences. The research ideas, methodology, experiments, and overall writing structure were entirely developed by the authors.

\subsection{Model}
\subsubsection{Single-agent perception pipeline}
\label{sec:single-agent-perception}
The perception pipeline comprises two components: a BEV encoder and a task-specific decoder. The BEV encoder projects sensor inputs into bird’s-eye-view (BEV) representations, enabling consistent spatial alignment and collaboration across different views. Task-specific decoders are then applied for downstream detection or segmentation tasks.

\textbf{BEV Encoder.}
Our framework supports either LiDAR or camera inputs, where we commonly denote the observation of the $i$th agent as $X_i$. We obtain BEV feature as $F_i=\Phi_{enc}(X_i)\in \mathbb{R}^{h \times w \times c}$, where $\Phi_{enc}(\cdot)$ denotes the complete BEV encoder. $F_i$ then serves as the information source for selection and compression.
For LiDAR inputs, we adopt the widely used PointPillars encoder~\cite{lang2019pointpillars} to extract BEV features from point clouds. For camera inputs, we employ the Lift-Splat-Shoot~\cite{philion2020lss} module following~\cite{lu2024HEAL}, which lifts image features into 3D frustums and aggregates them into the BEV plane via learned depth distributions. For both LiDAR and camera modality, the extracted BEV features are further processed by a 2D convolutional ResNeXt-based backbone~\cite{lu2024HEAL}.


\textbf{Decoder.}
Based on the BEV feature $F_i$, we incorporate task-specific decoders. 
For 3D object detection, the decoder consists of a classification head, a box regression head, and a direction estimation head to predict object bounding boxes, following~\cite{lu2024HEAL,hu2022where2comm}. 
For BEV semantic segmentation, we employ a MLP as decoder to produce dense, per-pixel semantic predictions, following~\cite{XuCoBEVT:CoRL22}.


\subsection{Theory}
\label{sec:proofs}
In this section, we provide: i) further discussion on our problem formulation; ii) proof of the proposed propositions and theories.

\subsubsection{Discussion on problem formulation}
\label{sec:discuss_problem_formulation}

\begin{wraptable}{r}{0.55\linewidth}
\centering
\caption{\small Problem formulations of bandwidth-constrained collaboration. $\mathcal{R}$: bit-rate, $\mathcal{D}$: distortion/loss.}
\label{tab:formulations}
\resizebox{\linewidth}{!}{
\begin{tabular}{lll}
\toprule    
\textbf{Formulation Type} & \textbf{Optimization target} & \textbf{Distortion/loss} $\mathcal{D}$ \\ \midrule
Constrained task loss~\cite{hu2022where2comm} & $\min \mathcal{D} \quad \text{s.t.}~ \mathcal{R} \leq \delta$  & task loss function \\
Task-compression joint loss~\cite{balle2018variational} & $\min \mathcal{D} + \lambda \mathcal{R}$  & task loss function \\
\textbf{ Pragmatic rate-distortion (ours)} & $\min \mathcal{R} \quad \text{s.t.}~ \mathcal{D} \leq \delta$ & pragmatic distortion 
\\ \bottomrule
\end{tabular}
}
\vspace{-2mm}
\end{wraptable}

Note that our theory formulation~\eqref{eq:collaboration_define}~\eqref{eq:collaboration_define_RD} is not an ad hoc assumption, but consistent with the learning objects in pragmatic compression~\cite{}. Table~\ref{tab:formulations} reveals that the optimization objective~\eqref{eq:collaboration_define_RD} is dual-equivalent to several prior approaches~\cite{hu2022where2comm,balle2018variational}, as they all share the same Lagrangian target $\min \mathcal{D} + \lambda \mathcal{R}$, which is a weighted sum of distortion $\mathcal{D}$ and communication rate $\mathcal{R}$ with weight $\lambda$. In the Section~\ref{sec:prag_distortion} we make the task distortion $\mathrm{D}_Y[X_s, Z_s \mid X_r]$ explicit. 

As an extreme case, consider early collaboration~\cite{han2023survey}, where agents directly transmit raw sensor data (i.e., $Z_s=X_s$). In this case, the communication volume becomes $\mathrm{I}(X_s;Z_s)=\mathrm{H}(X_s)$, corresponding to the full information content of $X_s$, and the distortion $\mathrm{D}_Y[X_s, X_s|X_r]$ is zero.

\subsubsection{Discussion on pragmatic distortion}

From Tab.~\ref{eq:distortion_det} we see that:  
i) for collaborative BEV segmentation, pragmatic distortion is expressed as the gap in conditional entropy, based on the Bayes risk \(\mathrm{B}_{risk}[Y|X] = \frac{1}{|Y|} \sum_{i \in \mathcal{S}} \mathrm{H}\left(Y_{(i)} | Z_s, X_r \right)\);  
ii) for collaborative 3D detection, the distortion further incorporates an exponential term of conditional entropy, where $\mathcal{K} = \{loc, size, ori\}$ denotes the set of regression losses (location, size, and orientation), which contribute more uncertainty than classification;  
iii) compared to the widely used MSE distortion in image reconstruction~\cite{balle2018variational}, the pragmatic distortion defined in our theory differs in two aspects: first, it emphasizes task uncertainty rather than fidelity, second, it accounts for the receiver’s information (e.g., $X_r$) to analyze redundancy.

\subsubsection{Proof of~\ref{pro:risk_centerpoint}: Bayes risk $\mathrm{R}[Y|X]$ for perception tasks}
\label{proof:risk_perception}

\begin{proposition}
    \label{pro:risk_centerpoint}
(Bayes risk $\mathrm{R}[Y|X]$ for perception tasks, see proof in~\ref{proof:risk_perception}). Given an observation input $X$ and a perception task target $Y$, we focus on the Bayes risk $\mathrm{R}[Y|X]$ to measure the difficulty of predicting $Y$ from $X$.
In object detection task, the detection results and corresponding target label are denoted as $\hat{Y}, Y\in \mathbb{R}^{h\times w\times (8+K)}$, where the $(8+K)$ channels stand for classification heatmap $\hat{Y}_{c},Y_{c}\in \mathbb{R}^{h\times w\times K}$, offset $\hat{Y}_{o},Y_{o}\in \mathbb{R}^{h\times w\times 3}$, size $\hat{Y}_{s},Y_{s}\in \mathbb{R}^{h\times w\times 3}$, rotation $\hat{Y}_{r},Y_{r}\in \mathbb{R}^{h\times w\times 2}$. Here $[h,w]$ denotes the BEV perception range. The total loss is shown in~\eqref{eq:loss_total_0}:

\vspace{-3mm}
\begin{align}
    L_{\text {total }}= L_{\text {heatmap }}+\lambda_2 L_{\text {offset }}+\lambda_3 L_{\text {size }}+\lambda_4 L_{\text {rotation }}
    \label{eq:loss_total_0}
\end{align}

where $\lambda_2,\lambda_3,\lambda_4$ are the loss weights. Consider $N$ objects involved in ground truth, the heatmap loss optimizes the model to classify the foreground object from background, where we adopt the focal loss $L_{\text {focal }}(y, \hat{y})=-\sum_{k=1}^K \alpha_k\left(1-\hat{y}_k\right)^\gamma y_k \log \hat{y}_k$, where $\alpha_k,\gamma$ are hyper-parameters in focal loss, here we consider a simplified situation that $\alpha_k=1,\gamma=0$ and the loss degenerates into cross entropy loss $L_{ce}$. $L_{\text {offset }}$,$L_{\text {size }}$, and $L_{\text {rotation }}$ are L1 loss.
Specifically, we consider the situation that the elements in $Y_o$ follow Gaussian distribution ${Y_o}_{(i,j)}| X \sim \mathcal{N}(\mu_o(X), \sigma_o^2(X))$, ${Y_s}_{(i,j)}| X \sim \mathcal{N}(\mu_s(X), \sigma_s^2(X)),{Y_r}_{(i,j)}| X \sim \mathcal{N}(\mu_r(X), \sigma_r^2(X))$, and the number of objects is $\overline{N_{obj}}$.
The Bayes risk of object detection with centerpoint detection loss is given as~\eqref{eq:risk_centerpoint_final_0}:

\vspace{-3mm}
\begin{align}
    \mathrm{R}[Y|X]&= \sum_{i\leq h,j\leq w} \mathrm{H}({Y_c}_{(i,j)} | X) + \overline{N_{obj}}\sqrt{2/ \pi}\left(\lambda_2 \sigma_o(X)  + \lambda_3 \sigma_s(X)  + \lambda_4\sigma_r(X)\right)
    \label{eq:risk_centerpoint_final_0}
\end{align}

When the task $Y$ refers to occupancy prediction, the regression terms are put off and the Bayes risk $\mathrm{R}[Y|X]$ solely consists of the terms of conditional entropy, as shown in~\eqref{eq:risk_occ_final_0}.
\vspace{-3mm}

\begin{align}
    \mathrm{R}[Y|X]&= \sum_{i\leq h,j\leq w} \mathrm{H}({Y_c}_{(i,j)} | X)= \sum_{i\leq h,j\leq w} \mathrm{H}({Y}_{(i,j)} | X)
    \label{eq:risk_occ_final_0}
\end{align}

\end{proposition}

In each communication round, messages are transmitted between connected agents as shown in \eqref{eq:collaboration_define}, where the connection is established by the pre-defined collaboration principle. We denote the observation of message sender/reciever as $X_s,X_r$, the perception target as $Y$. The message $\mathcal{P}_{s\rightarrow r}$ is obtained via $\mathcal{P}_{s\rightarrow r}=C(X_s)$, where $C(\cdot)$ is a compressor that reduces the transmission bit-rate.
The pragmatic distortion is defined in \eqref{eq:prag_distortion}, where $Y$ is the perception target, $\mathrm{R}[Y|X]$ denotes the Bayes risk when predicting $Y$ from $X$, $\mathrm{R}[Y|X_r,X_s]$ denotes the Bayes risk when predicting $Y$ from the fused information of $X_r,X_s$.

\vspace{-3mm}
\begin{align}
\mathrm{D}_Y\left[X_s, Z_s\right]=\mathrm{R}\left[Y \mid X_r, Z_s\right]-\mathrm{R}\left[Y \mid X_r, X_s\right]
\label{eq:prag_distortion}
\end{align}

To analyze this distortion in perception tasks, we need to:

\begin{itemize}
    \item Give the specific formulation of Bayes risk $\mathrm{R}[Y|X_r]$(single perception) and $\mathrm{R}[Y|X_r,X_s]$(collaborative perception) in detection 3D task with centerpoint loss(for example).
    \item Reformulate the distortion $\mathrm{D}_Y$ by introducing the task related Bayes risk $\mathrm{R}[Y|X_r,X_s]$.
    \item Reformulate the distortion $\mathrm{D}_Y$ by introducing the supply-request information.
\end{itemize}

\begin{definition}
\label{def:bayes_risk}
(Bayes risk). Let \( X \in \mathcal{X} \) be the input variable (features, observed data),  \( Y \in \mathcal{Y} \) be the target variable (labels), \( P(X, Y) \) denote the joint probability distribution of \( X \) and \( Y \), \( L(Y, \hat{Y}) \) be the loss function quantifying the discrepancy between a prediction \( \hat{Y} = f(X) \) and the true value \( Y \), and \( f: \mathcal{X} \to \mathcal{Y} \) be a predictive model.
The Bayes risk is defined as the infimum of the expected loss over all possible decision functions, as shown in \eqref{eq:bayes_risk}:

\vspace{-3mm}
\begin{align}
    \mathrm{R} = \inf_{f} \mathbb{E}_{X, Y} \left[ L(Y, f(X)) \right]
    \label{eq:bayes_risk}
\end{align}

Bayes risk is the minimum achievable loss by an ideally trained model. It captures unavoidable uncertainty in the data, such as the ambiguity due to overlapping classes in classification tasks or stochastic noise in target variables for regression tasks. For any model \( f \), the expected loss satisfies $\mathbb{E}[L(Y, f(X))] \geq \mathrm{R}_{\text{Bayes}}$. With a given loss function, the Bayes risk completely depends on the data distribution $P(X,Y)$, it indicates the "difficulty" of learning the projection $f: \mathcal{X} \to \mathcal{Y} $. Due to its property to characterize data distribution, we utilize the differences of Bayes risk to measure the pragmatic distortion.
\end{definition}

\subsubsubsection{Bayes risk for perception tasks}
In this section, we derive the Bayes risk of perception tasks with specific loss functions.

First, we review the formulation of centerpoint loss. Suppose that the observation from camera or LiDAR can be represented by 3D voxel feature $X \in \mathbb{R}^{D\times h\times w\times C}$, the detection results and corresponding target label are denoted as $\hat{Y}, Y\in \mathbb{R}^{h\times w\times (8+K)}$, where the $(8+K)$ channels stand for classification heatmap $\hat{Y}_{c},Y_{c}\in \mathbb{R}^{h\times w\times K}$, offset $\hat{Y}_{o},Y_{o}\in \mathbb{R}^{h\times w\times 3}$, size $\hat{Y}_{s},Y_{s}\in \mathbb{R}^{h\times w\times 3}$, rotation $\hat{Y}_{r},Y_{r}\in \mathbb{R}^{h\times w\times 2}$. The total loss is:

\vspace{-3mm}
\begin{align}
    L_{\text {total }}= L_{\text {heatmap }}+\lambda_2 L_{\text {offset }}+\lambda_3 L_{\text {size }}+\lambda_4 L_{\text {rotation }}
    \label{eq:loss_total}
\end{align}

where $,\lambda_2,\lambda_3,\lambda_4$ are the loss weights. Consider $N$ objects in ground truth, the heatmap loss optimizes the model to classify the foreground object from background, we utilize focal loss $L_{\text {focal }}(y, \hat{y})=-\sum_{k=1}^K \alpha_k\left(1-\hat{y}_k\right)^\gamma y_k \log \hat{y}_k$, where $\alpha_k,\gamma$ are hyper-parameters in focal loss, here we consider a simplified situation that $\alpha_k=1,\gamma=0$ and the loss degenerates into cross entropy loss $L_{ce}$. $L_{\text {offset }}$,$L_{\text {size }}$, and $L_{\text {rotation }}$ are L1 loss.
\newline

Now we derive the Bayes risk in 3D object detection with centerpooint detection loss. To simplify the formulation, we approximately decompose the total Bayes risk into the sum of Bayes risk on each location as shown in \eqref{eq:bayes_risk_decomposed}:
\vspace{-3mm}
\begin{align}
    \mathrm{R}[Y|X] = \inf_{f} \mathbb{E}_{X, Y} \left[ L(Y, f(X)) \right]=  \inf_{f} \sum_{i\leq h,j\leq w} \mathbb{E}_{X, Y_{(i,j)}} \left[ L(Y_{(i,j)}, f(X)_{(i,j)}) \right]
    \label{eq:bayes_risk_decomposed}
\end{align}

We regard the perception task at each region as independent tasks, and we define the located Bayes risk for perception tasks in~\eqref{eq:loc_bayes_risk}:

\vspace{-3mm}
\begin{align}
    \mathrm{R}[Y_{(i,j)}|X] =  \inf_{f} \mathbb{E}_{X, Y_{(i,j)}} \left[ L(Y_{(i,j)}, f(X)_{(i,j)}) \right]
    \label{eq:loc_bayes_risk}
\end{align}

We derive the Bayes risk for two primarily used loss function involved in perception tasks: focal loss, MSE loss, and L1 loss.

\textbf{Focal loss.} The true distribution of $P(Y_{c},X)$ satisfies $p({Y_c}_{(i,j,k)}=1 | X) = p_{i,j,k}$, and we have $P(Y_{c} | X) = \prod_{i,j,k} p({Y_c}_{(i,j,k)} | X)$ since different classes and locations are independent. The Bayes optimal prediction is the true conditional distribution: $\hat{{Y_c}}_{(i,j,k)}^* = p_{(i,j,k)}$, then we have:

\begin{align}
    \mathrm{R}_{\text{}}[{Y_c}_{(i,j)}|X]&= \mathbb{E}_{X,{Y_c}_{(i,j)}}  L_{ce}({Y_c}_{(i,j)},{p}_{(i,j)}) \\
    &= \mathbb{E}_{X} \sum_{{Y_c}_{(i,j)}} p({{Y_c}_{(i,j)}}|X)L_{ce}({Y_c}_{(i,j)},{p}_{(i,j)}) \\
    &= \mathbb{E}_{X} \sum_{k=1}^K -p({{Y_c}_{(i,j,k)}=1}|X)\log p({{Y_c}_{(i,j,k)}=1}|X) \\
    &= \mathrm{H}({Y_c}_{(i,j)} | X) 
\end{align}

\textbf{MSE loss.} Given a specific $X$, we assume that the elements in offset target $Y_o$ follow Gaussian distribution ${Y_o}_{(i,j)}| X \sim \mathcal{N}(\mu_o(X), \sigma_o^2(X))$, and the number of objects is $\overline{N_{obj}}$. This assumption is reasonable, since minimizing MSE loss can be regarded as MLE(Maximum likelihood estimation) when ${Y_o}_{(i,j)}| X  \sim \mathcal{N}(\mu_o(X), \sigma_o^2(X))$.
The Bayes optimal prediction is $f(X)=\hat{Y_o}_{(i,j)}^*=\mu_o(X)$. Put this into~\eqref{eq:bayes_risk_decomposed}, the Bayes risk is derived as $\overline{N_{obj}}\sqrt{2/ \pi} \sigma_o(X)$. Similarly, we can derive the Bayes risk for the size and rotation targets by assuming their distributions follow a Gaussian distribution ${Y_s}_{(i,j)}| X \sim \mathcal{N}(\mu_s(X), \sigma_s^2(X)),{Y_r}_{(i,j)}| X \sim \mathcal{N}(\mu_r(X), \sigma_r^2(X))$. Combining the Bayes risk of the individual loss function described in \eqref{eq:loss_total}, we obtain the Bayes risk of object detection with centerpoint detection loss as~\eqref{eq:risk_centerpoint_final}:

\vspace{-3mm}
\begin{align}
    \mathrm{R}_{centerpoint}[Y|X]&= \sum_{i\leq h,j\leq w} \mathrm{H}({Y_c}_{(i,j)} | X) + \overline{N_{obj}}\sqrt{2/ \pi}\left(\lambda_2 \sigma_o(X)  + \lambda_3 \sigma_s(X)  + \lambda_4\sigma_r(X)\right)
    \label{eq:risk_centerpoint_final}
\end{align}

\textbf{L1 loss.} Given a specific $X$, we assume that the elements in target $Y_o$ follow Laplace distribution $p(Y_o\mid X)=\frac{1}{2b_{o|X}}\exp\left(-\frac{|Y-\mu_o(X)|}{b_{o|X}}\right)$. This assumption is reasonable, since minimizing L1 loss can be regarded as MLE(Maximum likelihood estimation) when ${Y_o}_{(i,j)}| X  \sim \frac{1}{2b_{o|X}}\exp\left(-\frac{|Y-\mu_o(X)|}{b_{o|X}}\right)$. The Bayes optimal prediction is $f(X)=\hat{Y_o}_{(i,j)}^*=\mathrm{median}({Y_o}_{(i,j)}\mid X)=\mu_o(X)$. Put this into~\eqref{eq:bayes_risk_decomposed}, the Bayes risk is derived as:

\begin{align}
    \mathrm{R}_{\text{}}[{Y_o}_{(i,j)}|X]&= \mathbb{E}_{X,{Y_o}_{(i,j)}}  L_{l1}({Y_o}_{(i,j)},\mu_o(X))  && \hspace{-15mm} \text{definition} \\
    &= \mathbb{E}_{X} \int_{-\infty}^{\infty}|{Y_o}_{(i,j)}-\mu_o(X)|\cdot\frac{1}{2b_{o|X}}\exp\left(-\frac{|{Y_o}_{(i,j)}-\mu_o(X)|}{b_{o|X}}\right)d{Y_o}_{(i,j)} \\
    &= \mathbb{E}_{X} \frac{1}{2b_{o|X}}\int_{-\infty}^{\infty}|z|\exp\left(-\frac{|z|}{b_{o|X}}\right)dz && \hspace{-15mm} z={Y_o}_{(i,j)}-\mu_o(X) \\
    &= \mathbb{E}_{X} \frac{1}{2b_{o|X}} 2b_{o|X}^2 \\
    &= \mathbb{E}_{X} b_{o|X} \\
    &= b_{o|X} && \hspace{-15mm} b_{o|X}\text{ is a constant}
    \label{eq:risk_l1}
\end{align}

On the other hand, when $p({Y_o}_{(i,j)}=y| X=x) = \frac{1}{2b_{o|X}}\exp\left(-\frac{|y-\mu_o(x)|}{b_{o|X}}\right)$, we can formulate the conditional entropy $\mathrm{H}({Y_o}_{(i,j)}|X)$ as:

\begin{align}
    \mathrm{H}({Y_o}_{(i,j)}|X) &= \mathbb{E}_{x \sim X}  \mathrm{H}({Y_o}_{(i,j)}|X=x) \\
    &= \mathbb{E}_{x \sim X}\int_{-\infty}^{\infty}-p(y)\log p(y)\mathrm{~}dy \\
    &= \mathbb{E}_{x \sim X}\int_{-\infty}^{\infty}-\frac{1}{2b_{o|X}}\exp\left(-\frac{|y-\mu_o(x)|}{b_{o|X}}\right)(\log \frac{1}{2b_{o|X}}-\frac{|y-\mu_o(x)|}{b_{o|X}})\mathrm{~}dy \\
    &= \mathbb{E}_{x \sim X} \log (2b_{o|X})+1 \\
    &= \log (2b_{o|X})+1 && \hspace{-15mm} b_{o|X}\text{ is a constant}
    \label{eq:entropy_l1}
\end{align}

Combining~\eqref{eq:risk_l1} with~\eqref{eq:entropy_l1}, we have:
\begin{align}
    \mathrm{R}_{\text{}}[{Y_o}_{(i,j)}|X]=\frac{1}{2}\mathrm{e}^{\mathrm{H}({Y_o}_{(i,j)}|X)-1}
\end{align}

\subsubsection{Proof of Tab.~\ref{tab:distortions}: pragmatic distortion for collaborative perception.}
\label{proof:distor_coperception}

In this section, we derive the pragmatic distortion in collaborative perception.
To achieve this, we start from decomposing the contribution of ego agent and other agents.

Now we derive the pragmatic distortion in collaborative perception task. Consider a simple scenario with 2 collaborators, and the observations/features of message sender and receiver are $X_s$ and $X_r$, and the sender compresses $X_s$ into $Z_s$ to transmit, we define the pragmatic distortion as shown in~\eqref{eq:collab_distortion}, which measures the increase of Bayes risk after replacing the collaboration message $X_s$ with $Z_s$:

\vspace{-3mm}
\begin{align}
    \mathrm{D}_Y\left[X_s, Z_s\right]=\mathrm{R}\left[Y \mid X_r, Z_s\right]-\mathrm{R}\left[Y \mid X_r, X_s\right]
    \label{eq:collab_distortion}
\end{align}

We give a specific formulation by replacing the Bayes risk in ~\eqref{eq:collab_distortion} with  the Bayes risk of centerpoint loss in~\eqref{eq:risk_centerpoint_final}, as shown in~\eqref{eq:collab_distortion_centerpoint} :

\vspace{-3mm}
\begin{align}
    \label{eq:collab_distortion_centerpoint}
    \mathrm{D}_{Y_{(i,j)}}\left[X_s, Z_s\right]= \mathrm{H}({Y_c}_{(i,j)}|X_r,Z_s)-\mathrm{H}({Y_c}_{(i,j)}|X_r,X_s)+\\ \frac{1}{2}\lambda_2(\mathrm{e}^{\mathrm{H}({Y_o}_{(i,j)}|X_r,Z_s)-1}-\mathrm{e}^{\mathrm{H}({Y_o}_{(i,j)}|X_r,X_s)-1})+\\ \frac{1}{2}\lambda_3(\mathrm{e}^{\mathrm{H}({Y_s}_{(i,j)}|X_r,Z_s)-1}-\mathrm{e}^{\mathrm{H}({Y_s}_{(i,j)}|X_r,X_s)-1})+\\
    \frac{1}{2}\lambda_4(\mathrm{e}^{\mathrm{H}({Y_r}_{(i,j)}|X_r,Z_s)-1}-\mathrm{e}^{\mathrm{H}({Y_r}_{(i,j)}|X_r,X_s)-1})
    \label{eq:collab_distortion_detection}
\end{align}

We consider a degraded version by ignoring the regression loss, which is suitable for semantic occupancy prediction task, as shown in~\eqref{eq:collab_distortion_occupancy}:

\vspace{-3mm}
\begin{align}
    \label{eq:collab_distortion_occupancy}
    \mathrm{D}_{Y_{(i,j)}}\left[X_s, Z_s\right]= \mathrm{H}({Y}_{(i,j)}|X_r,Z_s)-\mathrm{H}({Y}_{(i,j)}|X_r,X_s)
\end{align}


\subsubsection{Proof of~\ref{theor:optimal_rate_collab}: optimal bit-rate of collaborative message}
\label{proof:collab_optimal_rate}

In this section, we derive the optimal transmission bit-rate in collaborative perception task. Consider the same collaboration situation described in Tab.~\ref{tab:distortions} with 2 collaborators, and the observations/features of message sender and receiver are $X_s$ and $X_r$, and the sender compresses $X_s$ into $Z_s$ to transmit. Our goal is to derive the minimum bit-rate needed to transmit $Z_s$ while guaranteeing a limited pragmatic distortion, as shown in~\eqref{eq:rate_distortion_collab}:

\vspace{-3mm}
\begin{align}
\operatorname{Rate}(\delta)=\min _{p(Z_s \mid X_s)} \mathrm{I}(X_s ; Z_s) \quad \text { s.t. } \mathrm{D}_Y[X_s, Z_s] \leq \delta .
\label{eq:rate_distortion_collab}
\end{align}

For occupancy prediction, put pragmatic distortion~\eqref{eq:collab_distortion_occupancy} into the constraint in~\eqref{eq:rate_distortion_collab}, we have~\eqref{eq:distortion_ineq}:

\vspace{-3mm}
\begin{align}
\mathrm{D}_Y[X_s, Z_s] &= \mathrm{H}({Y}|X_r,Z_s)-\mathrm{H}({Y}_{}|X_r,X_s) \\
&=[\mathrm{H}(Y|X_r)-\mathrm{I}(Y;Z_s|X_r)]-[\mathrm{H}(Y|X_r)-\mathrm{I}(Y;X_s|X_r)] \\
&=\mathrm{I}(Y;X_s|X_r)-\mathrm{I}(Y;Z_s|X_r) \leq  \delta
\label{eq:distortion_ineq}
\end{align}

This inequality condition also satisfies for object detection task, which corresponds to the distortion defined in~\eqref{eq:collab_distortion_detection} by considering two approximation:

\begin{itemize}
    \item First-order approximation.
\begin{align}
    \mathrm{e}^{\mathrm{H}({Y_o}_{(i,j)}|X_r,Z_s)-1}-\mathrm{e}^{\mathrm{H}({Y_o}_{(i,j)}|X_r,X_s)-1} \geq \frac{1}{\mathrm{e}}(\mathrm{H}({Y_o}_{(i,j)}|X_r,Z_s)-\mathrm{H}({Y_o}_{(i,j)}|X_r,X_s))
\end{align}
    \item Decomposition of joint entropy, with the assumption that the existing of object($Y_c$) is independent with the location ($Y_o$), size ($Y_s$), and heading ($Y_r$).
\begin{align}
    \mathrm{H}(Y_c,Y_o,Y_s,Y_r)=\mathrm{H}(Y_c)+\mathrm{H}(Y_o)+\mathrm{H}(Y_s)+\mathrm{H}(Y_r)
\end{align}
\end{itemize}

Given that, We reformulate ~\eqref{eq:rate_distortion_collab} as shown in~\eqref{eq:rate_distortion_collab_derive}:

\vspace{-3mm}
\begin{align}
\operatorname{Rate}(\delta)=&\min _{p(Z_s \mid X_s)~\text { s.t. } \mathrm{D}_Y[X_s, Z_s] \leq \delta } \mathrm{I}(X_s ; Z_s) \\
\geq &\min _{p(Z_s \mid X_s)~\text { s.t. } \mathrm{D}_Y[X_s, Z_s] \leq \delta } \mathrm{I}(X_s ; Z_s|X_r)
\label{eq:rate_distortion_collab_derive_0} \\
\geq & \min _{p(Z_s \mid X_s)~\text { s.t. } \mathrm{D}_Y[X_s, Z_s] \leq \delta } \mathrm{I}(Y ; Z_s|X_r)
\label{eq:rate_distortion_collab_derive_1} \\
\geq & \min _{p(Z_s \mid X_s)~\text { s.t. } \mathrm{D}_Y[X_s, Z_s] \leq \delta } \mathrm{I}(Y ; X_s|X_r)-\delta \\
=& \mathrm{I}(Y;X_s \mid X_r) -\delta \quad \hfill (\text{no} ~Z_s) \\
=& \mathrm{H}(X_s) - \Bigl[ \mathrm{H}(X_s) - \mathrm{I}(Y;X_s) \Bigr]  - \Bigl[ \mathrm{I}(Y;X_s) - \mathrm{I}(Y;X_s \mid X_r) \Bigr] -\delta \\
=& \mathrm{H}(X_s) - \underbrace{ \mathrm{H}(X_s|Y) }_{\text{information in }X_s\text{ irrelevant to }Y} - \underbrace{ \mathrm{I}(Y;X_s;X_r)}_{\text{information in }X_s\text{ redundant with }X_r\text{ about }Y} -\delta \quad \hfill
\label{eq:rate_distortion_collab_derive}
\end{align}

We make assumption that the variables follow the Markov chain $Y\leftrightarrow X_s \leftrightarrow Z_s$ and $X_r\leftrightarrow X_s \leftrightarrow Z_s$. Next, we will explain the reasoning behind each inequality and the conditions for these inequality to achieve equality.
\newline

The first inequality~\eqref{eq:rate_distortion_collab_derive_0} is satisfied when Markov chain $X_r\leftrightarrow X_s \leftrightarrow Z_s$ holds. This is because~\eqref{eq:conditional_MI_0}:
\vspace{-1mm}
\begin{align}
\mathrm{I}(Z_s;X_s,X_r)=\mathrm{I}(Z_s;X_r)+\mathrm{I}(Z_s;X_s|X_r)=\mathrm{I}(Z_s;X_s)+\mathrm{I}(Z_s;X_r|X_s)
\label{eq:conditional_MI_0}
\end{align}

\vspace{-1mm}
The Markov chain $X_r\leftrightarrow X_s \leftrightarrow Z_s$ leads to $\mathrm{I}(Z_s;X_r|X_s)=0$. Then we have~\eqref{eq:conditional_MI_1}:
\vspace{-1mm}
\begin{align}
\mathrm{I}(Z_s;X_s)=\mathrm{I}(Z_s;X_r)+\mathrm{I}(Z_s;X_s|X_r)\geq \mathrm{I}(Z_s;X_s|X_r)
\label{eq:conditional_MI_1}
\end{align}

\vspace{-1mm}
Here we can see the equality condition for the first inequality~\eqref{eq:rate_distortion_collab_derive_0} is that, $\mathrm{I}(Z_s;X_r)=0$, which means $Z_s$, the compressed version of $X_s$, should not have redundant information in $X_r$.
\newline

The second inequality~\eqref{eq:rate_distortion_collab_derive_1} is satisfied since due to DPI(Data Processing Inequality) given the Markov chain $Y\leftrightarrow X_s \leftrightarrow Z_s$. This is because~\eqref{eq:dpi_0}:
\vspace{-1mm}
\begin{align}
\mathrm{I}(Z_s;X_s,Y)=\mathrm{I}(Z_s;Y)+\mathrm{I}(Z_s;X_s|Y)=\mathrm{I}(Z_s;X_s)+\mathrm{I}(Z_s;Y|X_s)
\label{eq:dpi_0}
\end{align}

\vspace{-1mm}
The Markov chain $Y\leftrightarrow X_s \leftrightarrow Z_s$ leads to $\mathrm{I}(Z_s;Y|X_s)=0$. Then we have~\eqref{eq:dpi_1}:
\vspace{-1mm}
\begin{align}
\mathrm{I}(Z_s;X_s)=\mathrm{I}(Z_s;Y)+\mathrm{I}(Z_s;X_s|Y)\geq \mathrm{I}(Z_s;Y)
\label{eq:dpi_1}
\end{align}

\vspace{-1mm}
We can see that the equality condition for the second inequality~\eqref{eq:rate_distortion_collab_derive_1} is that, $\mathrm{I}(Z_s;X_s|Y)=0$. We can derive that~\eqref{eq:dpi_2}:
\vspace{-1mm}
\begin{align}
\mathrm{I}(Z_s;X_s|Y)=\mathrm{H}(Z_s|Y)-\mathrm{H}(Z_s|X_s,Y)=0
\label{eq:dpi_2}
\end{align}

\vspace{-1mm}
We can see from~\eqref{eq:dpi_2} that $\mathrm{H}(Z_s|Y)=\mathrm{H}(Z_s|X_s,Y)$, since $Z_s$ is a compressed version of $X_s$, the uncertainty $\mathrm{H}(Z_s|X_s,Y)$ is 0, therefore $\mathrm{H}(Z_s|Y)=0$. This implies that $Z_s$ is completely task-relative, it does not contains information unrelated to the task $Y$.
\newline

The third inequality is derived from~\eqref{eq:distortion_ineq}, and the equality condition is achieved when the distortion budget is sufficiently utilized.

\end{document}